\documentclass[preprint,3p,12pt]{elsarticle}
\usepackage[utf8]{inputenc}
\usepackage{amsmath,amssymb,amsfonts}
\usepackage{graphicx, enumitem}
\usepackage{textcomp}
\usepackage[ruled,vlined,linesnumbered]{algorithm2e}
\usepackage[table]{xcolor}
\usepackage{tikz}
\usepackage{bm}
\usepackage{booktabs}
\usepackage{multirow}
\usepackage{url}
\usepackage{caption}
\usepackage{subcaption}
\usepackage{tcolorbox}
\usepackage[left, modulo]{lineno}
\usepackage{comment}
\usepackage{courier}
\usepackage{tikz}
\usepackage{enumitem}
\usepackage{makecell}
\usepackage{times}
\usepackage{MnSymbol}

\usepackage{tcolorbox}

\SetKwInOut{Parameter}{Params}

\newtcbox{\mybox}{
  colback=teal,
  on line,
  before upper=\vphantom{ay},
}

\newcolumntype{L}[1]{>{\raggedright\let\newline\\\arraybackslash\hspace{0pt}}m{#1}}
\newcolumntype{C}[1]{>{\centering\let\newline\\\arraybackslash\hspace{0pt}}m{#1}}
\newcolumntype{R}[1]{>{\raggedleft\let\newline\\\arraybackslash\hspace{0pt}}m{#1}} 

\begin{document}
	
\begin{frontmatter}
	

\title{AiGAS-dEVL: An Adaptive Incremental Neural Gas Model for Drifting Data Streams under Extreme Verification Latency}
 
\author[mymainaddress]{Maria Arostegi\corref{mycorrespondingauthor}}
\ead{maria.arostegi@tecnalia.com}
\author[mythirdaddress]{Miren Nekane Bilbao}
\author[mymainaddress]{Jesus L. Lobo}
\author[mymainaddress,mythirdaddress]{Javier Del Ser}

\cortext[mycorrespondingauthor]{Corresponding author: TECNALIA. Parque Tecnologico, Ed. 700, 48160 Derio, Bizkaia, Spain. Telephone: (+34) 946 430 850. Fax: (+34) 901 760 009}

\address[mymainaddress]{TECNALIA, Basque Research and Technology Alliance (BRTA), 48160 Derio, Bizkaia, Spain}
\address[mythirdaddress]{University of the Basque Country (UPV/EHU), 48013 Bilbao, Spain}

\begin{abstract}
The ever-growing speed at which data are generated nowadays, together with the substantial cost of labeling processes (often reliant on human supervision) cause Machine Learning models to often face scenarios in which data are partially labeled, or delayed with respect to their query time. The extreme case where such a supervision is indefinitely unavailable is referred to as extreme verification latency. On the other hand, in streaming setups data flows are affected by exogenous factors that yield non-stationarities in the patterns to be modeled (concept drift), compelling models learned incrementally from the data streams to adapt their modeled knowledge to the prevailing concepts within the stream. In this work we address the casuistry in which these two conditions occur together over the data stream, by which adaptation mechanisms to accommodate drifts within the stream are challenged by the lack of supervision, hence requiring further mechanisms to track the evolution of concepts in the absence of verification. To this end we propose a novel approach, coined as AiGAS-dEVL (Adaptive Incremental neural GAS model for drifting Streams under Extreme Verification Latency), which relies on the use of growing neural gas to characterize the shape and inner point distributions of all concepts detected within the stream over time. Our approach exposes that the online analysis of the behavior of these prototypical points over time facilitates the definition of the evolution of concepts in the feature space, the detection of changes in their behavior, and the design of adaptation policies to mitigate the effect of such changes in the model. We assess the performance of AiGAS-dEVL over several synthetic datasets, comparing it to that of state-of-the-art approaches proposed in the recent past to tackle this stream learning setup. Our reported results reveal that AiGAS-dEVL performs competitively with respect to the rest of baselines, exhibiting a superior adaptability over several datasets in the benchmark while ensuring a simple and interpretable instance-based adaptation strategy.
\end{abstract}

	
\end{frontmatter}


\section{Introduction} \label{sec:intro}

The proliferation of data across diverse domains has undergone an unprecedented surge in recent years, driven by the increasing digitization of processes in almost all domains of activity, and the widespread proliferation of connected devices. This exponential growth in data generation is further compounded by the escalating scales and speeds at which data are being produced and disseminated on a daily basis. From social media interactions and online transactions to sensor readings in industrial IoT environments, the sheer volume and velocity of data streams pose grand challenges for traditional methods for batch data processing and analysis. Consequently, there is a compelling need to develop novel techniques and algorithms capable of handling the dynamic nature of data streams in real time. This growing concern has motivated the emergence of different research areas focused on the study of new efficient learning models for data analysis, including Big Data analytics \cite{kambatla2014trends}, real-time analytics \cite{ellis2014real} or, from a machine learning (ML) perspective, data stream learning \cite{gama2012survey,gomes2019machine}. When data are produced continuously, ingested and processed in real time, strict restrictions arise in terms of memory and processing power \cite{fan2013mining,krempl2011algorithm}, eventually constraining the design of ML algorithms operating over such data.

When addressing ML tasks formulated over data streams, the scalability of the learning algorithms is indeed a critical challenge, given the high volume and velocity of data being generated in a continuous fashion. Unfortunately, scalability is not the only challenge to be faced in such scenarios: there exist other issues that can significantly impact the effectiveness and reliability of ML models in this domain. On one hand, the concept drift phenomenon poses a major hurdle to the characterization of patterns within the data flows, as the underlying data distribution may change over time \cite{gomes2019machine,gama2014survey,ditzler2015learning,barcina2024managing}. When concept drift occurs, models trained on historical data become less accurate or even obsolete, unless their knowledge is modified and adapted to the prevailing data distribution. Hence, adaptation to evolving concepts becomes imperative for ensuring a sustained model performance under concept drift. On the other hand, extreme verification latency introduces another layer of complexity, where the delay in obtaining ground truth labels for incoming data can lead to discrepancies between the predictions issued by the model and their true values. This delay not only hinders model evaluation, but also exacerbates the adaptation to drifting data, as decisions need to be made in real-time based on potentially outdated or incomplete information. Consequently, addressing these challenges requires not only scalable learning algorithms, but also robust mechanisms for concept drift detection and adaptation, as well as strategies for handling extreme verification latency, ensuring the reliability and effectiveness of ML models over evolving data flows.

Interestingly under the scope of this work, most contributions reported to date around modeling over drifting data streams are focused on scenarios characterized by no latency, on the assumption of a so-called \emph{test-then-train} scheme. Consequently, the scope of such studies is restricted to improving the effectiveness of the adaptation strategies therein proposed assuming that the supervision of arriving data instances occurs immediately after they are predicted by the model at hand. Other works have instead addressed the case of a lagged label supervision with respect to the query time \cite{marrs2010impact}. Ultimately, when this verification latency is large or even infinite (\emph{extreme} as per the nomenclature used in the literature), an initial part of the data stream is assumed to be labeled, and after a certain point in time this supervision is no longer available. This models the situation when the speed at which stream data are generated make the supervision of data costly or even unfeasible in practice \cite{souza2015data}. When dealing with extreme verification latency \cite{razavi2019novelty}, semi-supervised and weakly supervised learning mechanisms have been investigated to exploit the unlabeled information arriving in the stream, especially when the initially labeled data does not suffice for reliably characterizing the distribution within the flowing data.

For years, researchers have reported advances and proposals for concept drift adaptation and learning over extreme verification latency independently, recognizing their impact on the efficacy of machine learning models operating over data streams. More recently, the simultaneous occurrence of both problems has garnered the attention of the community due to the higher modeling complexity produced by the convergence of these issues. While concept drift necessitates the continuous adjustment of models to evolving data distributions, extreme verification latency compounds the issue by introducing delays in feedback, hindering timely updates that can be informative and helpful for the detection and/or adaptation to the drift. Consequently, there is an interplay between concept drift and extreme verification latency by which the adaptation to drifts depend roughly on the capability of the model to trace and exploit past knowledge in an unsupervised manner. Indeed, among the different types of drift that may held in non-stationary data streams (in terms of intensity, duration or specific variable/relationship that evolves over time, e.g., concept- or feature-evolution \cite{khan2012tutorial}), the absence of immediate supervision makes the exploitation of knowledge captured by the model only feasible when dealing with \emph{slowly evolving} drifts (also referred to as \emph{gradual drifts}). Under these circumstances, concepts evolve over time progressively without sharp changes, enabling a chance for the model to track and adapt to the evolving patterns over time even in the lack of supervision.

Consequently, addressing both challenges concurrently becomes imperative for ensuring the reliability and effectiveness of ML models in dynamic data stream environments. Evidence of the growing interest in this paradigm are the different algorithmic proposals found in the recent literature to address together concept drift and extreme verification latency either separately or combined with each other. Our work aligns with this flurry of literature (later revised in the manuscript) by proposing a new adaptive ML approach to classification tasks formulated over gradually drifting data streams, capable of accommodating changes in the underlying concepts even in the absence of ground truth supervision. The proposal, coined as AiGAS-dEVL (Adaptive Incremental neural GAS model for drifting Streams under Extreme Verification Latency), hinges on the adoption of Growing Neural Gas (GNG) for the characterization of all concepts belonging to the classes existing in the stream, supporting their analysis and evolution tracking over time. This characterization, along with simple distance-based classification heuristics, provides an effective way for the prediction of unlabeled samples, which in turn helps towards the traceability of evolving concepts in the feature space. The key algorithmic contributions of this work can hence be summarized as follows:
\begin{itemize}[leftmargin=*]
\item We show that a continuous non-supervised characterization of incoming stream data by means of GNG can identify \textit{singular nodes} that determine the contour of the identified clusters. This characterization, alongside simple distance-based assignment heuristics, permits to \textit{trace} shape changes, \textit{trajectories} and \textit{varying number of concepts} in non-stationary streams in the lack of supervision. While the potential of instance-based adaptation mechanisms has been pinpointed in recent surveys \cite{suarez2023survey}, to the best of our knowledge there is no prior work prospecting this potential to yield specific algorithmic developments.

\item We devise a semi-supervised algorithm for the classification of singular nodes incrementally computed by GNG and newly incoming instances in the stream. It is important to notice at this point that the methodology proposed in this work expands beyond online clustering, since it is necessary to trace the correspondence between categories and concepts characterized and tracked over time. Hence, the overarching goal is to predict the classes of incoming stream instances using information on how data are organized into concepts that vary as a result of a gradual drift \cite{spiliopoulou2013monic}.
\end{itemize}

The performance of the proposed AiGAS-dEVL model is evaluated over a benchmark comprising several public datasets featuring evolving drifts of very diverse nature, assuming extreme verification latency in all of them. We also assess the performance over two real-world stream datasets used in the literature related to non-stationary data streams. Comparisons in terms of macro f1 score and prequential error against other baseline algorithms proposed for this same streaming setup expose that AiGAS-dEVL performs best in the benchmark, particularly when dealing with non-corpuscular concept shapes, varying number of concepts or when their evolution over time involve complex non-linear trajectories in the feature space. Our findings suggest that the unsupervised concept tracking mechanism embedded in AiGAS-dEVL opens new perspectives on the problem of unsupervised concept drift adaptation, and stimulates further follow-up studies outlined in the concluding section of the article.

The rest of the manuscript is organized as follows: Section \ref{sec:back} provides a brief literature review of stream learning, concept drift, extreme verification latency and GNG so as to place in context the contribution of this work. Next, Section \ref{sec:SSL-GNG} describes the proposed AiGAS-dEVL approach, including its general structure, detailed algorithmic steps, benefits and limitations. The experimental setup is detailed in Section \ref{sec:exp}, whereas results and comparisons to other baselines are presented and discussed in Section \ref{sec:exp_res}. Finally, Section \ref{sec:conc} ends the work by summarizing the main findings and sketching related research directions that can be pursued in the future.

\section{Background and Related Work} \label{sec:back}

Before proceeding with the detailed description of the proposed AiGAS-dEVL model, we first pause at the main milestones and advances attained recently in the fields of stream learning and concept drift (Section \ref{sec:SL_CD}), extreme verification latency (Section \ref{sec:EVL}) and GNG (Section \ref{sec:GNG}), establishing the grounds to justify the contribution of AiGAS-dEVL over the reviewed literature (Section \ref{sec:contrib}).

\subsection{Stream Learning and Concept Drift}\label{sec:SL_CD}

As has been argued in the introduction to this work, the significance of stream learning has grown exponentially in recent years, reflecting its pivotal role in addressing the computational and learning challenges posed by the rapid influx and dynamic nature of streaming data. Quickly arriving instances in a data stream can be retained only for a limited time budget, which impacts on the design of data processing pipelines, especially in hardware with constrained computing and storage capabilities \cite{stefanowski2017stream}. When the pipeline implies a ML model, constantly updating the knowledge captured by the model from the data stream is subject to the resources available in the device and the scales (volume and speed) at which data streams flow in the scenario at hand. As such, one can distinguish between \emph{on-line learning}, where the ML model is updated when a new instances arrives \textit{sequentially} from the stream, in a one-by-one fashion, whereas in incremental learning the model is updated over batches comprising a number of stream instances. 

Several comprehensive surveys on stream learning and concept drift have been published to date, emphasizing on the applications, challenges and taxonomies by which the plethora of contributions can be classified. Concept drift in data streams was formalized as early as 2009, capitalizing on prior studies dealing with changes in  the definitions of classes in learning tasks \cite{hand2006classifier}. Concept drift was later expanded by follow-up contributions discussing on the major challenges of data stream mining \cite{zliobaite2012next,krempl2014open,ditzler2015learning,lu2018learning}.

Concept drift refers to the phenomenon where the underlying data distribution changes over time, leading to shifts in the relationships between input features and target variables. Several types of concept drift can be distinguished, each with distinct characteristics and implications for model adaptation. First, a sudden (\emph{sharp}) concept drift occurs when the data distribution undergoes abrupt and significant changes, often resulting from unforeseen events or external factors. This type of drift poses a considerable challenge for ML models, as they must rapidly adjust to the new data distribution to maintain the level of performance achieved prior to the occurrence of the drift. In contrast, gradual concept drift involves more gradual and incremental changes in the data distribution over time, making it less immediately apparent but still requiring continuous model adaptation to ensure performance. Finally, recurring concept drift refers to patterns of change that occur periodically or intermittently, such as seasonal variations or cyclic trends, necessitating adaptive models capable of detecting and accommodating them effectively. Understanding the different types of concept drift is crucial for developing robust stream mining algorithms that can effectively handle the dynamic nature of streaming data \cite{vzliobaite2010learning}. 

Concept drift adaptation strategies in stream mining can be broadly categorized into active and passive approaches \cite{elwell2011incremental}. Active adaptation strategies involve actively monitoring data streams and triggering model updates or retraining when concept drift is detected. Examples of active approaches include drift detection algorithms that continuously analyze incoming data for deviations from the established model, prompting timely adjustments to ensure model accuracy. On the other hand, passive adaptation strategies involve building models that inherently adapt to changes in the data distribution without explicit drift detection mechanisms. These strategies often involve the use of dynamic learning algorithms, ensemble methods, or online learning techniques that incrementally update model parameters as new data arrives, allowing the model to adapt gradually to evolving concepts. While active approaches offer real-time responsiveness to concept drift, they may incur higher computational costs and require more sophisticated drift detection mechanisms. In contrast, passive approaches provide a more seamless and resource-efficient adaptation process, but can be less sensitive to sudden or unexpected changes in the data distribution. 

\subsection{Extreme Verification Latency}\label{sec:EVL}

In scenarios subject to sparse annotation \cite{korycki2022instance} or ultimately, extreme verification latency, the scarcity or absence of incoming labeled data samples requires the use of different adaptation strategies to overcome eventual changes in the distribution of data \cite{siekmann2007knowledge,marrs2010impact,umer2017learning}. In scenarios where ground truth labels are not immediately available after prediction, ML models operating over data streams must contend with the uncertainty introduced by the lag between the arrival of a data instance and the verification of its true label. Without timely feedback, ML models may become outdated or misaligned with evolving data distributions, compromising their performance and reliability. Adaptation mechanisms become essential in such environments to enable ML models to dynamically adjust their parameters and update their internal representations based on the limited and delayed feedback received. Ultimately, when the supervision delay is infinite, adaptation becomes even more complicated due to the lack of information signaling whether the adaptation was effective in its purpose.   

Among the different methods to handle verification latency, our proposed AiGAS-dEVL approach connects closely with semi-supervised learning approaches \cite{gomes2019machine,li2019incremental}. In this family of models, a few labeled samples are available for initially model training, which allows subsequently extracting further knowledge from the large amount of unsupervised streaming data. Differently from semi-supervised learning, active learning methods \cite{settles2012active}, select the instances to be learned by the model. Given the computational constraints posed by stream learning and the evolving nature assumed for the data streams, active learning can be thought to be the most suitable approach for the scenario under hypothesis \cite{gomes2019machine}. AiGAS-dEVL also selects which data points to retain in its base knowledge towards prediction, hence it can be regarded as an active semi-supervised learning technique.

Several algorithms published in recent years tackle classification tasks formulated over non-stationary data streams subject to extreme verification latency. To begin with, the so-called Arbitrary Sub-Populations Tracker (APT) \cite{krempl2011algorithm} utilizes expectation-maximization to determine the optimal one-to-one assignment between the unlabeled and labeled data (by using kernel density estimation techniques), so that the classifier is subsequently updated after unlabeled data are labeled. Another renowned semi-supervised technique is COMPOSE (Compacted Object Sample Extraction) \cite{dyer2013compose}, which proposes a geometry-based framework to learn from non-stationary streaming data. COMPOSE follows three steps: 1) $\alpha$-shapes that represent the current class conditional distribution are constructed; 2) $\alpha$-shapes are compacted (shrinked) to represent the geometric center of each class distribution; and 3) from the compacted $\alpha$-shapes new instances are extracted to serve as labeled data for future time steps. 

A further step was taken by the so-called Stream Classification Algorithm Guided by Clustering (SCARGC) proposed in \cite{souza2015data}. SCARGC harnesses current and past cluster positions obtained by clustering unlabeled data to track drifting classes over time. Contemporaneously, evolutions of COMPOSE were contributed to improve different aspects, including the computational requirements and speed with respect to the seminal proposal of this semi-supervised learner. This is the case of FAST COMPOSE \cite{umer2016learning}, which overrides the need for extracting the most representative samples of the feature space of every class, yielding a more efficient approach for high-dimensional settings; Modular Adaptive Sensor System (MASS) \cite{frederickson2017adding}, a derivative of COMPOSE that uses centroid-based clustering and an update rule to maintain the computed centroids updated over time; and LEVELIW (Learning Extreme VErification Latency with Importance Weighting) \cite{umer2017level}, which iteratively weights the importance of chosen instances using those labeled in the previous time step, feeding them to an importance weighted least squares probabilistic classifier.

We end our short review on the related literature by commenting on AMANDA \cite{ferreira2019amanda}, a semi-supervised density-based adaptive model for non-stationary data for weighting and filtering samples that best represent the concepts in evolving distributions. Its density-based nature enables AMANDA to adapt dynamically to changes in data, ensuring its effectiveness even in the presence of shifting patterns in the stream. AMANDA comprises two different variants: AMANDA-FCP and AMANDA-DCP, the latter overcoming the strong dependence of the approach on one of its hyper-parameters (the cutting percentage). Shortly before, TRACE \cite{arostegi2018concept} showcased the inherent utility of short-term trajectory prediction for tracking the evolution of concepts in gradual drifts in an unsupervised manner. In a follow-up work we proposed SLAYER (Concept Tracking and Adaptation for Drifting Data Streams under Extreme Verification Latency) \cite{arostegi2021slayer}, an algorithm that extended TRACE with mini-batch incremental clustering and distance-based matching to reach a better adaptability to drifting streams with varying number of concepts.

\subsection{Growing Neural Gas}\label{sec:GNG}

A proper understanding of Growing Neural Gas (GNG) departs from Self Organizing Maps (SOMs), a neural computation method to construct representations from high-dimensional data. Unlike other dimensionality reduction approaches, SOMs generate a map of similarity relationships. That is, SOMs learn topological relations based on a distance metric. In doing so, SOMs are trained using unsupervised learning and a \textit{competitive} strategy, producing a discrete representation of the input space in a lower-dimensional map. The dimension of the input space also sets the number of input neurons each being connected to all the output neurons (of the map). These connections do not quantify the importance of each of the neurons, but translate the position it will end up occupying in the map.

Unlike SOM, where the topology of the network (number of units and connections) is predefined, GNG learns the topology by adding both units and connections as more information is received. That is, the Growing Neural Gas (GNG) \cite{sun2017online} is an incremental neural model able to learn the topological relations of a given set of patterns by means of competitive Hebbian learning. The first work in this direction appears on 1994, \cite{fritzke1994growing}, which introduces this model as an incremental network capable to learn topological relationships within a given set of input vectors by means of a simple Hebbian-like learning rule. Since then, the community has proposed optimization techniques aimed to speed up the learning algorithm of the naive GNG model \cite{fivser2013growing}. More recent research works underscore the potential of GNG to provide an overview of the structure of large volumes of multidimensional data, which can be used in exploratory data analysis \cite{ventocilla2019visual,ventocilla2021scaling}. 

\subsection{Contribution}\label{sec:contrib}

This last perspective on the use of GNGs settles the starting hypothesis of the present work: since self-organizing neural networks try to preserve the topology of an input space through competitive learning \cite{bouguelia2018adaptive}, we hypothesize that this capability could be exploited for capturing the evolving representation of data instances over time. These models, particularly GNG, can  contribute to the characterization and tracking of the impact of gradual concept drifts on the feature space and the concepts underneath the flowing data, even in the absence of annotation. This is indeed the motivation for embracing GNG at the core of the AiGAS-dEVL model proposed in this work: the use of GNG not only allows maintaining a continuously updated characterization of the data stream, but is furthermore endowed with additional functionalities such that, if a concept drift is detected, the model adapts itself accordingly. These functionalities make AiGAS-dEVL a passive approach suited to deal with gradual concept drifts.

\section{Proposed AiGAS-dEVL Approach}\label{sec:SSL-GNG}

In this section we delve into the algorithmic elements comprising the proposed AiGAS-dEVL model, including the mathematical definition of the dEVL problem (Subsection \ref{sec:mathematical_notation}) and a detailed description of the proposed algorithm and its steps (Subsection \ref{sec:General_Scheme}). We end by analyzing the benefits and points of improvement of AiGAS-dEVL in Subsection \ref{sec:complexity_adv_lim}.

\subsection{Problem Statement} \label{sec:mathematical_notation}

An dEVL problem is defined on a stream of data instances $\{(\mathbf{x}_t,y_t)\}_{t=1}^\infty$, where $\mathbf{x}_t\in\mathbb{R}^N$ denotes the data instance flowing at time $t$ and $y_t\in\mathcal{Y}$ its label. We assume that labels are available for $t<T_s$, so that a model $M_{\bm{\theta}}(\mathbf{x})$ with parameters $\bm{\theta}$ can be learned over the samples during this supervised period. For $t>T_s$, labels of the instances arriving from the stream are not available any longer, so the learned model is used to predict them as $\widehat{y}_t = M_{\bm{\theta}}(\mathbf{x}_t)$ for $t\geq T_s$. The challenge emerges when the distribution $p(y|\mathbf{x})$ characterized by $M_{\bm{\theta}}(\mathbf{x})$ is non-stationary, i.e., $p_t(y|\mathbf{x})\neq p_{t'}(y|\mathbf{x})$ for $t,t'\geq T_s:\: t\neq t'$. The goal is to endow the learned model $M_{\bm{\theta}}(\cdot)$ with the capability to adjust its modeled knowledge to the changes (also referred to as \emph{concept drift}) undergone by the stream data distribution over time. The lack of supervision for $t\geq T_s$ requires that the drift dynamics are gradual, so that the evolving concepts can be \emph{tracked} by the augmented model in the absence of supervision, i.e. without having access to $\{y_t\}_{t\geq T_s}$. We note that this definition also applies to the case of mixed drift, i.e., when both $p(\mathbf{x})$ and $p(y|\mathbf{x})$ change over time. This last case affects both the distribution of features and the decision boundary of the model, requiring more sophisticated adaptation mechanisms to adapt its knowledge.

\subsection{Description of AiGAS-dEVL}\label{sec:General_Scheme}

The design of our devised AiGAS-dEVL algorithm departs from the estimation of one of the key parameters of the approach from the supervised part of the data stream: the initial number of nodes characterized by GNG. For the sake of computational efficiency, AiGAS-dEVL utilizes a single GNG algorithm to characterize $p_t(\mathbf{x})$ over time. In the case of a very unbalanced dataset, stream data instances belonging to the most prevalent classes can dominate the production of nodes by GNG, leaving the other classes underrepresented in the node distribution output by GNG. To overcome this, the initial number of GNG nodes is adjusted as $G_0 = (1+\xi)G$, where $G$ is an hiper-parameter and $\xi$ is given by the ratio of the number of samples of the most to the least populated class in $\{\mathbf{x}_t,y_t\}_{t<T_s}$. This parameter is used by GNG to learn an initial map of nodes $\{\mathbf{x}_g^{\boxplus,0}\}_{g=1}^{G_0}$ over $\{\mathbf{x}_t,y_t\}_{t<T_s}$. Such nodes can be assigned a label $y_g^{\boxplus,0}\in\mathcal{Y}$ by means of a classifier $M_{\bm{\theta},ini}(\cdot)$ learned from $\{\mathbf{x}_t,y_t\}_{t<T_s}$, i.e. $y_g^{\boxplus,0} = M_{\bm{\theta},ini}(\mathbf{x}_g^{\boxplus,0})$. While our experiments detailed later consider a $K_{\textup{GNG}}$-Nearest Neighbor algorithm $\textup{NN}(\cdot)$ as $M_{\bm{\theta},ini}(\cdot)$, any other classifier can be used instead.

Once this initial supervised period is over, extreme verification latency occurs, hence only unsupervised instances $\{\mathbf{x}_t\}_{t\geq T_s}$ are received. We assume that instances are received in \emph{batches}, so that $\{\mathbf{x}_t\}_{t\geq T_s}$ is redefined as $\{\mathbf{x}^b\}_{b=1}^\infty$, where $b$ is the batch index and $\mathbf{x}^b\doteq[\mathbf{x}_1^b,\ldots,\mathbf{x}_B^b]$, with $B$ denoting the batch size. Labels of instances within the first unsupervised batch $b=1$ arriving from the stream are predicted by measuring their distance to the already computed GNG nodes $\{\mathbf{x}_g^{\boxplus,0}\}_{g=1}^{G_0}$, and voting the assigned labels of the $K$ nearest GNG nodes to every instance in the batch. Moreover, the GNG is updated with this first batch of stream instances, producing a new set of GNG nodes $\{\mathbf{x}_g^{\boxplus,1}\}_{g=1}^{G_1}$. Such nodes are annotated by searching for the $K_{\textup{GNG}}$ nearest neighbors of every node \smash{$\mathbf{x}_g^{\boxplus,1}$} in the initial set of nodes \smash{$\{\mathbf{x}_g^{\boxplus,0}\}_{g=1}^{G_0}$}. In this way, the updated distribution of GNG nodes is ready to predict the unsupervised stream instances received in the next batch.
\begin{figure}
	\centering
	\includegraphics[width=\columnwidth]{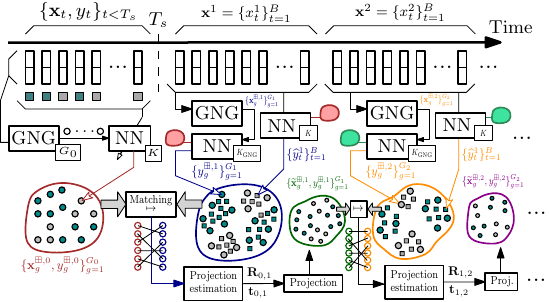}
	\caption{General diagram of the algorithmic flow followed by AiGAS-dEVL. GNG stands for \emph{Growing Neural Gas}, and NN for \emph{Nearest Neighbors} model. Symbols $\circ$ refer to GNG nodes, whereas $\square$ denote the data instances arriving from the stream in batches $\mathbf{x}^b$, with $b\in\{1,\ldots,\infty\}$.}
	\label{fig:general}
\end{figure}

At this point it is important to note that given the assumption of drifting data during the EVL period, the distribution of nodes $\{\mathbf{x}_g^{\boxplus,1}\}_{g=1}^{G_1}$ and their assigned supervision $\{y_g^{\boxplus,1}\}_{g=1}^{G_1}$ can be refined to better predict instances in the next batch by estimating the change dynamics of the distribution of nodes from $b-1$ to $b$, so that GNG nodes estimated from batch $b$ can be \emph{projected} following the same dynamics. This projection idea is summarized graphically in Figure \ref{fig:general}, alongside the rest of algorithmic steps comprised by AiGAS-dEVL. As can be observed in this figure, the premise that the drift suffered by $p_t(\mathbf{x})$ and/or $p_t(y|\mathbf{x})$ occurs gradually over time allows for the characterization of such dynamics based on a two-fold process:
\begin{enumerate}[leftmargin=*]
\item Solving a minimum-cost assignment problem to map all nodes in $\{\mathbf{x}_g^{\boxplus,1}\}_{g=1}^{G_1}$ to a node from the previous distribution $\{\mathbf{x}_g^{\boxplus,0}\}_{g=1}^{G_0}$. Here, cost is given by the $G_0\times G_1$ distance matrix $\mathbf{D}_{0,1}=[d_{g',g'}]$, such that the goal is to find a $G_0\times G_1$ binary matrix $\mathbf{A}_{0,1}=[a_{g,g'}]$ as:
\begin{equation}\label{eq:minimum_cost_problem}
\begin{aligned}
	&\min_{\mathbf{A}_{0,1}} && \sum_{g=1}^{G_0} \sum_{g'=1}^{G_1} d_{g,g'} a_{g,g'}, \\
	&\textup{subject to:} &&\sum_{g=1}^{G_0} a_{g,g'}=1\quad \forall g'\in\{1,\ldots,G_1\}, \\
	&&&\sum_{g'=1}^{G_1} a_{g,g'}\leq 1\quad \forall g'\in\{1,\ldots,G_1\}.
\end{aligned}
\end{equation}

The above formulation is an instance of the so-called \emph{rectangular} assignment problem, for which several efficient solving algorithms have been proposed in the literature \cite{bijsterbosch2010solving}. In the case of our designed AiGAS-dEVL approach, we embrace the polynomial-time modified Jonker-Volgenant solver proposed in \cite{crouse2016implementing}, whose efficiency and scalability are suitable for the computational requirements of stream learning.

\item Estimating a projection based on the estimated correspondence between $\{\mathbf{x}_g^{\boxplus,0}\}_{g=1}^{G_0}$ and $\{\mathbf{x}_g^{\boxplus,1}\}_{g=1}^{G_1}$ through $\mathbf{A}_{0,1}$. Among all the possible projections, given the progressive drift dynamics of the stream we opt for a simple rigid transformation given by $\mathbf{t}\in\mathbb{R}^{N}$ (translation vector) and $\mathbf{R}\in\mathbb{R}^{N\times N}$ (rotation matrix). Such matrices can be estimated by virtue of several decomposition algorithms; among them, AiGAS-dEVL seeks the optimal rotation and translation $\mathbf{R}^\ast$ and $\mathbf{t}^\ast$ such that:
\begin{equation} \label{eq:rotation_translation}
\mathbf{R}^\ast,\:\mathbf{t}^\ast = \arg \min_{\mathbf{R},\mathbf{t}} \sum_{g'=1}^{G_1} \sum_{g=1}^{G_0} a_{g,g'}\cdot ||\mathbf{R}\cdot \left(\mathbf{x}_g^{\boxplus,0}+\mathbf{t}\right)-\mathbf{x}_{g'}^{\boxplus,1}||^2,
\end{equation} 
namely, a least-square fitting of $\mathbf{R}$ and $\mathbf{t}$ considering the mapping $\mathbf{A}_{0,1}$. This problem can be solved by the well-known Kabsch algorithm \cite{kabsch1976solution}, also referred to as the Kabsch-Umeyama algorithm \cite{umeyama1991least}, which also meets the computational constraints imposed by stream learning by virtue of its $\mathcal{O}(G_1)$ complexity.
\end{enumerate}

Once this projection estimation is done, the nodes $\{\mathbf{x}_g^{\boxplus,1}\}_{g=1}^{G_1}$ computed for batch $b=1$ can be projected through the estimated $\mathbf{R}$ and $\mathbf{T}$, yielding:
\begin{equation} \label{eq:projection}
\widetilde{\mathbf{x}}_{g}^{\boxplus,1} = \mathbf{R}\cdot (\mathbf{\mathbf{x}_g^{\boxplus,1}} + \mathbf{t}), \quad \forall g\in\{1,\ldots,G_1\},
\end{equation}
with its assigned label being that of the node from which it is projected, i.e., $y_g^{\boxplus,1}$. This above set of projected nodes serves as the basis for the prediction of the next batch of samples $\{\mathbf{x}_b^2\}_{b=1}^B\doteq \mathbf{x}^2$ (i.e. $b=2$) arriving at the stream. In doing so, the $K$ nearest neighbors of every sample $\mathbf{x}_b^2$ to the distribution of projected nodes $\widetilde{\mathbf{x}}_{g}^{\boxplus,1}\}_{g=1}^{G_1}$ is utilized, so that their predicted label is given by voting the annotations $y_g^{\boxplus,1}$ associated to such neighboring nodes.

Subsequently, the steps to process and predict the rest of batches in the stream follow analogously: 
\begin{enumerate}[leftmargin=*]
	\item Labels associated to the samples within a new batch $\mathbf{x}^b$ are predicted based on the $K$ nearest neighbors of the projected nodes  $\{\widetilde{\mathbf{x}}_{g}^{\boxplus,b-1}\}_{g=1}^{G_{b-1}}$ of batch $b-1$.
	\item The distribution of nodes is updated via GNG and their labels are assigned based on the $K_{\textup{GNG}}$ neighbors of every node $\mathbf{x}_g^{\boxplus,b}$ to $\{\mathbf{x}_g^{\boxplus,b-1}\}_{g=1}^{G_{b-1}}$.
	\item The mapping $\mathbf{A}_{b-1,b}$ between the new node distribution $\{\mathbf{x}_g^{\boxplus,b},y_g^{\boxplus,b}\}_{g=1}^{G_b}$ and $\{\mathbf{x}_g^{\boxplus,b-1}\}_{g=1}^{G_{b-1}}$ is computed as per Expression \eqref{eq:minimum_cost_problem}.
	\item Rotation matrix $\mathbf{R}_{b-1,b}$ and translation vector $\mathbf{t}_{b-1,b}$ are estimated\footnote{The notation is slightly changed to explicitly reflect the dependence of $\mathbf{R}_{b-1,b}$ and $\mathbf{t}_{b-1,b}$ on the batch indices $b-1$ and $b$ of the mapped GNG node distributions.} based on Expression \eqref{eq:rotation_translation}.
	\item A new projected distribution of GNG nodes $\{\widetilde{\mathbf{x}}_{g}^{\boxplus,b}\}_{g=1}^{G_b}$ is obtained by applying Expression \eqref{eq:projection} for $g\in\{1,\ldots,G_{b}\}$, so that steps 1 to 4 can be iterated over a new arriving batch $\mathbf{x}^{b+1}$.
\end{enumerate}
\begin{algorithm}[ht]
	\caption{Proposed AiGAS-dEVL algorithm.}\label{alg:SSL_GNG}
	\DontPrintSemicolon
	\SetAlgoLined
	\SetKwInOut{Input}{Input}
	\SetKwInOut{Output}{Output}	
	\Input{Supervised data instances $\{\mathbf{x}_t,y_t\}_{t<T_s}$, unsupervised stream instances $\{\mathbf{x}_t\}_{t\geq T_s}$, batch size $B$, initial number of GNG nodes $G$.}	
	\Output{Predicted labels $\{\widehat{y}_t\}_{t\geq T_s}$.}
	Compute the proportion of supervised instances for each class.\;
	Let $G_0=(1+\xi)G$, where $\xi$ is the ratio between the maximum and minimum proportion of supervised instances for every class.\;
	Learn a model $M_{\bm{\theta},ini}(\cdot)=NN(\cdot;K,\{\mathbf{x}_t,y_t\}_{t<T_s})$.\;
	Compute an initial distribution of nodes \smash{$\{\mathbf{x}_g^{\boxplus,0}\}_{g=1}^{G_0}$} via GNG over \smash{$\{\mathbf{x}_t\}_{t<T_s}$}.\;
	Predict nodes' labels \smash{$\{y_g^{\boxplus,0}\}_{g=1}^{G_0}$} as \smash{$y_g^{\boxplus,0}=\textup{NN}(\mathbf{x}_g^{\boxplus,0};\{\mathbf{x}_t\}_{t<T_s},K)$} for $g\in\{1,\ldots,G_0\}$.\;
	Let \smash{$\widetilde{\mathbf{x}}_g^{\boxplus,0}=\mathbf{x}_g^{\boxplus,0}$} $\forall g\in\{1,\ldots,G_0\}$.\;
	\For{$b\in[1,\ldots,\infty]$}
	{
		Collect a new batch of streaming data instances $\mathbf{x}^b\doteq [\mathbf{x}_1^b,\ldots,\mathbf{x}_B^b]$.\;
		Predict their labels as \smash{$\widehat{y}_t^b=\textup{NN}(\mathbf{x}_t^b;\{\widetilde{\mathbf{x}}_g^{\boxplus,b-1}\}_{g=1}^{G_{b-1}},K)$} for $t\in\{1,\ldots,B\}$.\;
		Compute GNG nodes of the new batch as \smash{$\{\mathbf{x}_g^{\boxplus,1}\}_{g=1}^{G_1}$}\;
		Predict their labels as \smash{$y_g^{\boxplus,b}=\textup{NN}(\mathbf{x}_g^{\boxplus,b};\{\widetilde{\mathbf{x}}_g^{\boxplus,b-1}\}_{g=1}^{G_{b-1}},K_\textup{GNG})$} for $g\in\{1,\ldots,G_b\}$.\;
		Calculate $\mathbf{A}_{b-1,b}$ between \smash{$\{\mathbf{x}_g^{\boxplus,b-1}\}_{g=1}^{G_{b-1}}$} and \smash{$\{\mathbf{x}_g^{\boxplus,b}\}_{g=1}^{G_b}$} as per \eqref{eq:minimum_cost_problem}.\;
		Compute the rotation matrix $\mathbf{R}_{b-1,b}$ and translation vector $\mathbf{t}_{b-1,b}$ as per \eqref{eq:rotation_translation}.\;
		Compute projected GNG nodes \smash{$\{\widetilde{\mathbf{x}}_g^{\boxplus,b}\}_{g=1}^{G_{b}}$} using $\mathbf{R}_{b-1,b}$ and $\mathbf{t}_{b-1,b}$ as per \eqref{eq:projection}.\;
	}
	The predicted labels are given by the collated predictions for every batch (line \textbf{8}).
\end{algorithm}


Algorithm \ref{alg:SSL_GNG} summarizes the main steps of the proposed AiGAS-dEVL approach, comprising the input parameters, the expected outputs (namely, the labels of the unsupervised part of the stream), and the different steps performed during the supervised (lines \textbf{1} to \textbf{6}) and unsupervised (lines \textbf{7} to \textbf{15}) parts of the stream. The former essentially computes a $G_0$-sized distribution of GNG nodes based on the samples belonging to the supervised set of stream instances (lines \textbf{1} to \textbf{4}), and predicts their labels based on the $K$ nearest neighbors within such supervised stream data instances (line \textbf{5}). Then, imposing that no projection is computed at the beginning of the algorithm (line \textbf{6}), AiGNG-dEVL iterates over the batches successively received over time by collecting the data instances, predicting them based on the prevailing (projected) distribution of GNG nodes (line \textbf{9}) and calculating the new distribution of nodes and their labels (lines \textbf{10} and \textbf{11}). Then, for every batch the projection best relating the previous and current node distributions is estimated (line \textbf{13}) via minimum-distance matching between nodes (line \textbf{12}). Finally, the distribution of nodes is projected accordingly, yielding an updated set of prevalent GNG nodes that can be used for predicting the next batch of collected data instances considering the drift dynamics of the stream (line \textbf{14}).

\subsection{Benefits and Limitations} \label{sec:complexity_adv_lim}

Before proceeding with the experimental discussion of this manuscript, we first examine and reflect on the benefits and points of improvement of AiGAS-dEVL:
\begin{itemize}[leftmargin=*]
\item Benefits:
\begin{itemize}[leftmargin=*]
\item \emph{Computational complexity}: For every batch, AiGAS-dEVL comprises several processing steps that contribute to its overall computational effort during the unsupervised part of the stream. They include NN inference (lines \textbf{9} and \textbf{11}), GNG update (line \textbf{10}), minimum-distance matching (line \textbf{12}) and the estimation of the parameters driving the assumed rigid drift dynamics (line \textbf{13}). On the one hand, efficient implementations of the neighbor search algorithm exist, leveraging ball tree sorting mechanisms that can lower down the overall search complexity per query from $\mathcal{O}(1)$ (training) and $\mathcal{O}(BNK)$ (inference) to $\mathcal{O}(BN\log(B))$ and $\mathcal{O}(BN)$, respectively. Since the two NN inferences per batch are applied on the same training data $\{\widetilde{x}_g^{\boxplus,b-1}\}_{g=1}^{G_{b-1}}$, the sorting procedure can be implemented only once per batch. When it comes to the GNG update, its complexity in batch $b\in\{1,\ldots,\infty\}$ is given by $\mathcal{O}(G_{b-1})$. As mentioned previously, a modified Jonker-Volgenant solver \cite{crouse2016implementing} is used to compute the matching between the previous and current GNG node distributions, bounding its complexity as $\mathcal{O}(G_b^3)$. Finally, the computation of $\mathbf{R}_{b-1,b}$ and $\mathbf{t}_{b-1,b}$ that models how the streaming data distribution evolves over time has a linear complexity $\mathcal{O}(G_b)$.

\item \emph{Interpretability}: the use of GNG provides further benefits beyond modeling the distribution of the streaming data over time. Together with its distance-based prediction criteria and the assumption of rigid drift dynamics, AiGAS-dEVL can be considered to fall within the family of algorithmically transparent models, namely, the algorithm in its seminal form can be understood and even replicated by an user without requiring a deep knowledge on Machine Learning or data processing. Indeed, GNG is a relatively simple and straightforward algorithm, without complex mathematical formulations or opaque internal mechanisms. The core steps of the algorithm -- finding the nearest neurons, updating their positions, and adding new neurons -- are easy to understand and visualize. Further along this line, the topology produced by GNG directly reflects the structure of the input data, allowing for the direct observation how the algorithm models the input data using 2D or 3D visualizations of the network.

\item \emph{Adaptability to evolving feature spaces}: GNG is not only well-suited for streaming data, but it also has the ability to adapt to non-corpuscular topologies effectively. One of the key advantages of GNG is its ability to dynamically grow and adapt its network structure to the underlying data distribution, without requiring any prior assumptions about the shape or topology of the data. Unlike many other clustering or dimensionality reduction algorithms that work best with convex, well-separated clusters, GNG can handle complex, non-convex, and even non-corpuscular data structures. Furthermore, the lack of strict assumptions about the data topology enables GNG to be effectively applied to streaming data scenarios, where the data distribution may evolve over time. 

\item \emph{Implicit forgetting mechanisms}: since GNG is designed to continuously adapt to input data, it suitable adapts to changing data distributions over time. While this adaptability does not constitute a forgetting mechanism per se, it does enable GNG to adjust to new patterns in streaming data. This is accomplished not only by adding new neurons and connections to the network, but also removing old edges. Moreover, the matching between previous and new GNG node distributions (line \textbf{12} in Algorithm 1) provides an implicit forgetting mechanism, as not all nodes from batch $b-1$ must be mapped to those produced for batch $b$.

\item \emph{Design flexibility}: when it regards to the ML method producing the sought predictions $\{\widehat{y}_t^b\}_{t=1}^B$ $\forall b$, AiGAS-dEVL can be regarded as a model-agnostic method that can seamlessly accommodate any other approach different than $\textup{NN}(\cdot)$ for predicting such labels. A similar statement holds when it comes to the estimation of the labels corresponding to the new distribution of nodes estimated by GNG for every new batch $b$ in the stream (line \textbf{11} in Algorithm 1). Likewise, other projections can be considered to infer the drift undergone by the stream in its unsupervised regime, capable of characterizing more sophisticated change dynamics in the feature space. Therefore, AiGAS-dEVL can be thought of to be a data processing and modeling framework that allows for the consideration of manifold combinations of methods under the same design principles.

\item Lack of complex hyperparameters: GNG has relatively few hyper-parameters to tune, such as the learning rates and the maximum number of neurons. This simplicity contributes to the overall transparency of the algorithm. Furthermore, techniques to automate the GNG configuration process in streaming environments have been reported in the recent literature (e.g. Bayesian optimization \cite{zhang2024bayes}), which can be also integrated into the overall workflow of AiGAS-dEVL.
\end{itemize}

\item Limitations and points of improvement:
\begin{itemize}[leftmargin=*]
\item \emph{Dependence on the drift dynamics}: the projection chosen to model the evolution of concepts between consecutive batches may not match the drift dynamics of the stream. Our description of AiGAS-dEVL assumed a rigid transformation for this projection, which does not suffice to model non-linear warps in the evolution of GNG concepts. Nevertheless, the flexibility of AiGAS-dEVL permits to overcome this shortcoming by using new functions for this transformation, alongside techniques to estimate their parameters based on $\{\mathbf{x}_g^{\boxplus,b-1}\}_{b=1}^{G_{b-1}}$ and  $\{\mathbf{x}_g^{\boxplus,b}\}_{b=1}^{G_{b}}$, e.g. polynomial warping or thin plate spline transformations.

\item \emph{Lack of memory in the assignment of labels to new GNG nodes}: the distribution of GNG nodes $\{\mathbf{x}_g^{\boxplus,b}\}_{g=1}^{G_b}$ corresponding to every incoming batch $\mathbf{x}^b$ is annotated based on the annotation of the GNG distribution of the previous batch of samples $\{y_g^{\boxplus,b-1}\}_{g=1}^{G_{b-1}}$. In the presence of concepts belonging to different classes, the changes induced by the drift can make them coexist in the same region of the feature space. If such concepts collide with each other for several consecutive batches, it can be problematic for properly tracking the correspondence between concepts and labels if the algorithmic step in charge of this assignment is not designed to remember the correspondence prior to the collision between concepts. This can eventually give rise to a mismatch between concepts and labels that, in the absence of supervision, can propagate catastrophically over the rest of the stream without any chance for recovering it.
\end{itemize}
\end{itemize}

Our experiments consider the seminal version of AiGAS-dEVL described in Algorithm 1, which suffice for providing a competitive performance when compared to other state-of-the-art variants for dEVL scenarios. However, these points of improvement will be part of our future research lines, which we will further detail in the prospects closing the conclusions of the manuscript.

\section{Experimental Setup}\label{sec:exp}

In order to assess the performance of the proposed AiGAS-dEVL algorithm, several experiments are carried out to provide empirical evidence to the answers to two research questions (RQ):
\begin{itemize}[leftmargin=*]
\item RQ1: \textit{How does AiGAS-dEVL perform when compared to other approaches proposed for modeling drifting data streams with extreme verification latency?}
\item RQ2: \textit{Do the points of improvement identified previously reflect on the performance of AiGAS-dEVL over time under different drift dynamics?}
\end{itemize}

\paragraph{Datasets} Our designed experimental setup considers a public repository of non-stationary data streams widely adopted by the stream mining community \cite{souza2015data}. Specifically, the repository contains $15$ synthetic datasets featuring gradual drift changes over time, and $2$ real datasets. Table \ref{tab:datasets} summarizes their main characteristics, where \emph{drift interval} refers to the interval in number of examples between consecutive drifts, and $N$ and $|\mathcal{Y}|$ refer to the number of features and classes, respectively. A column labeled as \emph{Drift} is added to the table to specify whether the drift dynamics of every dataset are \emph{rectilinear}. This aspect will be of relevance when discussing on the results obtained to address RQ2. 

The first real dataset is \texttt{ELEC2}, which has become a typical benchmark dataset in streaming data classification. It was first described in \cite{harries1999splice} and used thereafter for several performance comparisons \cite{baena2006early,kuncheva2008adaptive}. It contains data collected by the Australian New South Wales (NSW) Electricity Market, amounting to a total of $45,312$ instances. Once normalized and cleaned, $27,552$ instances dated from May 1996 to December 1998 are considered. Each example of the dataset has $N=5$ attributes: the day of the week, the time stamp, the NSW electricity demand, the Vic electricity demand and the scheduled electricity transfer between states. The class label represents the change of the price ($\mathcal{Y}=\{\textup{UP},\textup{DOWN}\}$) in New South Wales relative to a moving average of the last 24 hours. The distribution of classes in this dataset is not balanced, and the drift time is unknown. 
\begin{table}[h!]
\centering
\caption{Datasets for non-stationary data stream classification under EVL \cite{souza2015data} and real-world datasets considered in the experiments.}
\label{tab:datasets}
\begin{tabular}{cccccc}
\toprule
\textbf{Dataset} & $\bm{|\mathcal{Y}|}$ & $\bm{N}$ & \textbf{\# instances} & \textbf{Drift interval} & \textbf{Drift} \\ \midrule
\texttt{1CDT}           & 2       & 2        & 16,000         & 400  & Rectilinear \\ 
\texttt{1CHT}           & 2        & 2        & 16,000        & 400 &  Rectilinear \\
\texttt{2CDT}            & 2        & 2        & 16,000         & 400 &  Rectilinear \\
\texttt{2CHT}            & 2        & 2          & 16,000       & 400 & Rectilinear \\
\texttt{4CR}           & 4           & 2         & 144,400         & 400 &  Rectilinear \\
\texttt{4CRE-V1}       & 4           & 2          & 125,000        & 1,000 &  -\\ 
\texttt{4CRE-V2}       & 4             & 2        & 183,000         & 1,000 &  -\\ 
\texttt{5CVT}          & 5             & 2        & 40,000          & 1,000 & Rectilinear \\  
\texttt{1CSURR}          & 2        & 2        & 55,283        & 600  &  - \\
\texttt{4CE1CF}        & 5          & 2         & 173,250         & 750 & Rectilinear  \\ 
\texttt{UG2C2D}   & 2          & 2           & 100,000        & 1,000 & -  \\ 
\texttt{MG2C2D}    & 2         & 2          & 200,000        & 2,000  &  -  \\ 
\texttt{FG2C2D}    & 2         & 2           & 200,000      & 2,000 &  -  \\ 
\texttt{UG2C5D}    & 2          & 5         & 200,000         & 2,000 &  -  \\ 
\texttt{GEARS}   & 2         & 2           & 200,000       & 2,000 &  -  \\ 
\midrule
\texttt{ELEC2}    & 2           & 8          & 45,312         & Unknown &  -  \\  
\texttt{KEYSTROKE}   & 4          & 10         & 1,600         & 200 &  -  \\ 
\bottomrule
\end{tabular}
\end{table}

The second real dataset is \texttt{KEYSTROKE}, which yields from the use of keystroke dynamics to identify users by their typing rhythm instead of simply relying on username and password verification. The dataset, which aims to classify the users, contains $N=10$ features extracted from the time interval between releasing a key and pressing the next one \cite{souza2015classification}. In this streaming dataset, four users were chosen randomly, to merge them later respecting the chronological order, obtaining at the end of the process a total of $1,600$ instances. This dataset is balanced and the drift occurs around the $200$-th instance.

\paragraph{Baselines} Several algorithms from the literature related to EVL in the presence of concept drift (reviewed in Subsection \ref{sec:EVL}) are considered in our experiments:
\begin{itemize}[leftmargin=*]
\item  A static classifier (\textbf{STC}), learned from the first labeled samples and kept fixed over time (i.e., it does not adapt its knowledge).

\item  A sliding window classifier (\textbf{SLD}), which learns initially from the labeled samples and updates its knowledge with predicted upcoming samples, discarding those instances that do not fall inside the sliding window for predicting new instances.

\item An incremental window classifier (\textbf{INC}), which is similar to the sliding window classifier, but does not forget any past instance for training.

\item COMPOSE (\textbf{CMP}, \cite{dyer2013compose}), which resorts to a core of samples from the current data and defines a shape that represents the distribution of each class. When a new batch arrives, COMPOSE extracts instances from the core region(s) to use them as training data. In this manner, the new unlabeled data instances are combined with the core to adapt the knowledge of the classifier to non-stationarities occurring in the stream.

\item LEVELiw (\textbf{LVL}, \cite{umer2016learning,umer2017level}), which was created for data stream in EVL scenarios undergoing slow concept drifts. It uses an iterative weighting technique as long as an overlap between conditional class distributions between consecutive time steps prevails.

\item AMANDA \cite{ferreira2019amanda}, which leverages labeled samples to define the distribution of each class, and trains a semi-supervised classifier to classify the unlabeled batches arriving from the stream. To improve its accuracy, AMANDA uses density-based mechanisms to measure the importance of the classified instances, weighting them and keeping the most representative ones. This method has two variants: AMANDA-FCP (\textbf{A-FCP}), which selects a fixed number of samples; and AMANDA-DCP (\textbf{A-DCP}), which dynamically selects samples from the data stream.
\end{itemize} 

\paragraph{Configuration} In accordance with common practice in the area, we assume a batch scenario in which $5\%$ of the dataset is assumed to be labeled ($t<T_s$), considering the remaining samples ($t\geq T_s$) as unlabeled. The unsupervised part is divided in $100$ batches arriving in chronological order. The size of the batch $B$ yields from this configuration and the length of the unsupervised part of every dataset. We also set $M_{mb}=20$ mini-batches per batch for \textbf{SLAYER}. The hyper-parameter values of all baselines are kept to the same used in \cite{ferreira2019amanda}. As for our approach, $G$ is chosen to be $100$ nodes for all datasets. 

\paragraph{Evaluation} The performance of the above baselines and the proposed AiGAS-dEVL model over the datasets listed in Table \ref{tab:datasets} is gauged in terms of the \emph{prequential error}, which is a referential measure in the data stream literature \cite{gama2009issues,gama2013evaluating}. The prequential error allows monitoring the evolution of the performance of models over time. In order to use the same metrics as in \cite{ferreira2019amanda}, the prequential error is computed based on an accumulated sum of a loss function between the predicted and observed values, i.e.:
\begin{equation}
P_e (t)=\frac{1}{t} \sum_{t'=1}^{t}\mathcal{L}(y_{t'},\widehat{y}_{t'}), 
\end{equation}
where the prequential error is computed at time $t$, $\widehat{y}_{t'}$ represents the prediction at time $t'$, and $y_{t'}$ represents its ground truth value. The loss function $\mathcal{L}(\cdot,\cdot)$ is set to the zero-one loss, i.e., for every stream instance the loss equals 1 if it is wrongly predicted, and 0 otherwise. In addition to this performance score, we also report the average macro F1 scores of the models computed over every dataset in a non-prequential fashion. This measure is given by:
\begin{equation}
F_1 = \frac{2 \cdot \text{Precision} \cdot \text{Recall}}{\text{Precision} + \text{Recall}}
\end{equation}
i.e., the harmonic mean of precision and recall averaged over all classes and the entire set of unsupervised samples of every dataset. Performance statistics (mean, standard deviation) are provided across all datasets for every approach, whereas the statistical significance is assessed by means of a Bayesian analysis of the differences \cite{benavoli2017time} between every pair of algorithms in the benchmark. The analysis of the significance focuses on the top performing baselines, and considers varying values of the \emph{rope} parameter used to define the range of difference values in which the algorithms under comparison are considered to be equivalent.

The source code implementing the proposed approach and the results of the comparison study presented in what follows is available in the Github repository \url{https://git.code.tecnalia.com/maria.arostegi/aigas-devl}.

\section{Results and Discussion}\label{sec:exp_res}

We now present and discuss on the results obtained for every research questions formulated above:

\subsection*{RQ1: How does AiGAS-dEVL perform when compared to other approaches proposed for modeling drifting data streams with extreme verification latency?}
 
The results obtained for this first research question are summarized in Table \ref{tab:results} (average prequential error) and Table \ref{tab:f1} (average macro F1 score). In these tables, the best and second best counterparts for every dataset are highlighted as \fcolorbox{teal!30}{teal!30}{\vphantom{\parbox[c]{1cm}{\rule{1cm}{0.2cm}}}$\quad$} and \fcolorbox{gray!20}{gray!20}{\vphantom{\parbox[c]{1cm}{\rule{1cm}{0.2cm}}}$\quad$}, respectively. The last two rows inform about the mean and standard deviation statistics of every algorithm computed over all datasets. The statistical significance of the gaps is discussed later.
\begin{table}[h!]
	\centering
	\caption{Prequential error of the compared methods for the considered datasets.}
	\renewcommand{\arraystretch}{0.92}
	\resizebox{\columnwidth}{!}{
		\begin{tabular}{cccccccccc}
			\toprule
			\textbf{Dataset} & \textbf{STC} & \textbf{SLD} & \textbf{INC} & \textbf{CMP} & \textbf{LVL} & \textbf{A-FCP} & \textbf{A-DCP} & \textbf{SLAYER} & \makecell{\textbf{AiGAS-dEVL}\\(proposed)} \\
			\midrule
			\texttt{1CDT} & 0.76 & 0.06 & 0.3 & 0.08 & 0.04 & \cellcolor{gray!20}0.02 & 0.05 & \cellcolor{teal!30} 0.013 & 0.04 \\
			\texttt{1CHT} & 3.93 & 0.43 & 3.2 & 0.48 & 0.4 & \cellcolor{teal!30}0.33 & 0.39 & 0.4 & \cellcolor{gray!20}0.37 \\
			\texttt{2CDT} & 46.3 & 6.13 & 46.14 & 6.73 & 49.74 & \cellcolor{gray!20}5.46 & 5.83 & \cellcolor{teal!30}3.8 & 8.6
			\\
			\texttt{2CHT} & 45.97 & 48.45 & 46.01 & 47.41 & 47.41 & \cellcolor{gray!20}14.39 & 19.93 & \cellcolor{teal!30}10.6 & 15.15 \\
			\texttt{4CRT} & 78.83 & \cellcolor{gray!20}0.02 & 78.75 & 0.04 & \cellcolor{gray!20}0.02 & \cellcolor{gray!20}0.02 & 0.03 & \cellcolor{teal!30}0.01 & \cellcolor{teal!30}0.01 \\
			\texttt{4CRE-V1} & 78.15 & 81.29 & 79.44 & 79.55 & 79 & 73.5 & 73.28 & \cellcolor{teal!30}2.64 & \cellcolor{gray!20}3.1 \\
			\texttt{4CRE-V2} & 79.61 & 82.88 & 79.67 & 77.38 & 80.77 & 69.97 & 71.81 & \cellcolor{gray!20}8.06 & \cellcolor{teal!30}8.03 \\
			\texttt{5CVT} & 54.51 & 60.97 & 52.04 & 65.5 & 59.18 & 24.11 & 52.38 & \cellcolor{gray!20}21.9 & \cellcolor{teal!30}20.22 \\
			\texttt{1CSURR} & 35.86 & 9.05 & 36.06 & 9.43 & 9.2 & \cellcolor{gray!20}4.39 & 7.93 & 15.58 & \cellcolor{teal!30}3.8
			\\
			\texttt{4CE1CF} & 1.98 & 1.9 &  \cellcolor{gray!20}1.82 & 2.09 & 2.21 & \cellcolor{teal!30}1.73 & 1.92 & 2.1 & 2.05
			\\
			\texttt{UG2C2D} & 55.81 & 4.97 & 54.42 & 5.32 & 26.34 &  \cellcolor{gray!20}4.3 & 12.64 & 4.51 & \cellcolor{teal!30}4.28 \\
			\texttt{MG2C2D} & 51.63 & 22.86 & 50.66 & 49.17 & 9.31 & \cellcolor{gray!20}8.7 & 14.88 & 14.91 & \cellcolor{teal!30}7.38 \\
			\texttt{FG2C2D} & 17.79 &  \cellcolor{gray!20}4.43 & 18.29 & 12.15 & \cellcolor{teal!30}4.31 & 5.12 & 16.39 & 16.25 & 4.5 \\
			\texttt{UG2C5D} & 30.97 & 20.11 & 30.62 & 20.82 & 20.82 & \cellcolor{gray!20}8.21 & 8.53 & 8.31 & \cellcolor{teal!30}7.73 \\
			\texttt{GEARS} & 5.43 &  \cellcolor{gray!20}0.81 & 5.33 & 4.03 & 6.18 &  \cellcolor{gray!20}0.81 & 3.74 & 4.2 & \cellcolor{teal!30}0.43
			\\
			\texttt{ELEC2} & 29.22 & 41.04 & 28.92 & 36.31 & 32.64 &  \cellcolor{teal!30}26.12 & 34.45 & 42.2 &  \cellcolor{gray!20}26.3 \\
			\texttt{KEYSTROKE} & 32.3 & 12.37 & 12.5 & 14.67 & 12.97 & \cellcolor{gray!20}9.87 & \cellcolor{teal!30}9.34 & 49.4 & 11.9 \\
			\midrule
			Average & 38.18 & 23.40 & 36.72 & 25.36 & 25.91 & 15.12 & 19.62 &  \cellcolor{gray!20}12.05 & \cellcolor{teal!30}7.29 \\
			Std. Dev. & 26.69 & 28.69 & 27.42 & 28.10 & 27.58 & 22.67 & 24.03 &  \cellcolor{gray!20}14.30 & \cellcolor{teal!30}7.45 \\
			\bottomrule
		\end{tabular}
	}
	\label{tab:results}
\end{table}

We begin our discussion by analyzing the prequential error statistics in Table \ref{tab:results}. As expected due to the non-stationarity of the stream datasets under consideration, methods not designed to cope with this circumstance (namely, \textbf{STC}, \textbf{SLD} and \textbf{INC}) yield in general comparatively poor results. However, a closer inspection of the results reveals that in three datasets (\texttt{GEARS}, \texttt{4CRT}, \texttt{FG2C2D} and \texttt{4CEF1CF}), the prequential error of these naive methods gets closer to that of the rest of algorithms in the benchmark. Conversely, the results of baselines that encompass adaptive mechanisms to deal with concept drifts under unsupervised stream data regimes, such as \textbf{CMP} or \textbf{LVL}, show an improvement when compared to the aforementioned naive methods. Nevertheless, they still lag behind those obtained by \textbf{A-FCP}, \textbf{A-DCP}, \textbf{SLAYER} and the proposed \textbf{AiGAS-dEVL}. The statistics at the bottom of the table evince that our proposal performs best on average across the datasets under consideration, yielding in addition a lower variability of the prequential error. 
\begin{table*}[!h]
	\centering
	\vspace{-3mm}
	\caption{Average macro F1 results of the compared methods for the considered datasets.}
	\label{tab:f1}
	\resizebox{\textwidth}{!}{
		\begin{tabular}{cccccccccc}
			\toprule
			\textbf{Dataset} & \textbf{STC} & \textbf{SLD} & \textbf{INC} & \textbf{CMP} & \textbf{LVL} & \textbf{A-FCP} & \textbf{A-DCP} & \textbf{SLAYER} & \makecell{\textbf{AiGAS-dEVL}\\(proposed)}  \\
			\midrule
			\texttt{1CDT} & 0.994 & \cellcolor{gray!20}0.999 & 0.997 & \cellcolor{teal!30}1.000 & \cellcolor{teal!30}1.000 & \cellcolor{teal!30}1.000 & \cellcolor{gray!20}0.999 & \cellcolor{teal!30}1.000 & \cellcolor{gray!20}0.999 \\
			\texttt{1CHT} & 0.960 & 0.995 & 0.968 & 0.995 & \cellcolor{gray!20}0.996 & \cellcolor{gray!20}0.996 & \cellcolor{gray!20}0.996 & \cellcolor{teal!30}0.997 & \cellcolor{gray!20}0.996 \\
			\texttt{2CDT} & 0.387 & \cellcolor{gray!20}0.942 & 0.388 & 0.936 & 0.484 & \cellcolor{teal!30}0.948 & \cellcolor{gray!20}0.942 & 0.933 & 0.912 \\
			\texttt{2CHT} & 0.395 & 0.356 & 0.394 & 0.476 & 0.476 & \cellcolor{gray!20}0.853 & 0.788 & \cellcolor{teal!30}0.962 & 0.844 \\
			\texttt{4CR} & 0.210 & \cellcolor{teal!30}1.000 & 0.215 & \cellcolor{teal!30}1.000 & \cellcolor{teal!30}1.000 & \cellcolor{teal!30}1.000 & \cellcolor{teal!30}1.000 & \cellcolor{gray!20}0.999 & \cellcolor{teal!30}1.000 \\
			\texttt{4CRE-V1} & 0.207 & 0.180 & 0.200 & 0.204 & 0.249 & 0.267 & 0.265 & \cellcolor{gray!20}0.967 & \cellcolor{teal!30}0.968 \\
			\texttt{4CRE-V2} & 0.204 & 0.126 & 0.204 & 0.197 & 0.246 & 0.304 & 0.181 & \cellcolor{gray!20}0.910 & \cellcolor{teal!30}0.919 \\
			\texttt{5CVT} & 0.354 & 0.181 & 0.371 & 0.239 & 0.177 & 0.730 & 0.380 & \cellcolor{gray!20}0.789 & \cellcolor{teal!30}0.808 \\
			\texttt{1CSURR} & 0.640 & 0.914 & 0.638 & 0.909 & 0.637 & \cellcolor{teal!30}0.961 & 0.927 & 0.857 & \cellcolor{gray!20}0.960 \\
			\texttt{4CE1CF} & \cellcolor{gray!20}0.981 & 0.980 & \cellcolor{teal!30}0.982 & 0.978 & 0.978 & \cellcolor{gray!20}0.981 & 0.980 & 0.974 & 0.978 \\
			\texttt{UG2C2D} & 0.443 & 0.951 & 0.455 & 0.949 & 0.737 & \cellcolor{teal!30}0.958 & 0.871 & 0.954 & \cellcolor{gray!20}0.957 \\
			\texttt{MG2C2D} & 0.480 & 0.754 & 0.494 & 0.505 & 0.592 & \cellcolor{gray!20}0.914 & 0.850 & 0.850 & \cellcolor{teal!30}0.923 \\
			\texttt{FG2C2D} & 0.732 & \cellcolor{gray!20}0.939 & 0.730 & 0.860 & \cellcolor{teal!30}0.947 & 0.932 & 0.819 & 0.810 & \cellcolor{gray!20}0.939 \\
			\texttt{UG2C5D} & 0.668 & 0.755 & 0.678 & 0.792 & 0.792 & \cellcolor{gray!20}0.915 & 0.913 & 0.913 & \cellcolor{teal!30}0.922 \\
			\texttt{GEARS} & 0.947 & \cellcolor{teal!30}0.996 & 0.949 & 0.964 & 0.938 & \cellcolor{teal!30}0.996 & 0.963 & 0.957 & \cellcolor{gray!20}0.994 \\
			\texttt{ELEC2} & 0.617 & 0.374 & 0.610 & 0.623 & 0.447 & \cellcolor{gray!20}0.690 & 0.555 & 0.360 & \cellcolor{teal!30}0.714 \\
			\texttt{KEYSTROKE} & 0.727 & 0.952 & \cellcolor{gray!20}0.956 & 0.850 & 0.788 & \cellcolor{teal!30}0.983 & 0.935 & 0.540 & 0.880 \\
   \midrule
   Average & 0.585 & 0.729 & 0.602 & 0.734 & 0.675 & 0.849 & 0.786 & \cellcolor{gray!20}0.869 & \cellcolor{teal!30}0.924 \\
   Std. Dev. & 0.278 & 0.336 & 0.290 & 0.298 & 0.288 & 0.230 & 0.269 & \cellcolor{gray!20}0.173 & \cellcolor{teal!30}0.077 \\
			\bottomrule
	\end{tabular}}
\end{table*}

When turning the scope of our discussion to the F1 scores collected in Table \ref{tab:f1}, similar conclusions can be drawn. While non-adaptive techniques do perform well in some datasets, in general the benchmark is dominated by those including adaptation funcionalities in their algorithmic design. Remarkably, a wider spread is observed in terms of ranking, yet performance differences between the best approaches for every dataset are narrower than in the case of the prequential error. There emerges the need for examining qualitatively whether there is any relationship between the drift dynamics of the datasets and the performance behavior of the algorithms in the benchmark. This is indeed the motivation of RQ2, which will be later addressed in our discussions. In regards to the quantitative analysis made in response to RQ1, the narrower performance gaps require further statistical analysis to shed light on the significance of the performance differences in Tables \ref{tab:results} and \ref{tab:f1}.

\begin{figure}[h!]
	\centering
	\subfloat[]{{\includegraphics[width=0.23\columnwidth]{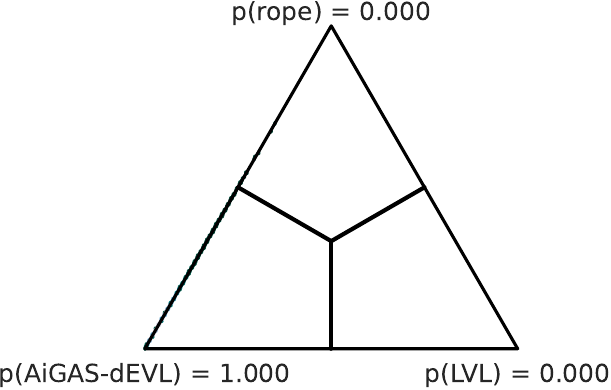}}}%
	\hfill
	\subfloat[]{{\includegraphics[width=0.23\columnwidth]{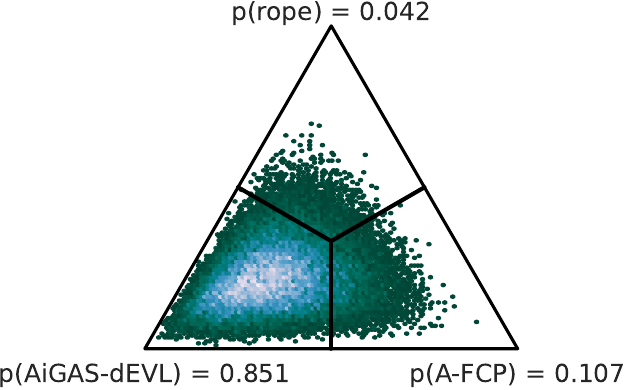}}}%
	\hfill
	\subfloat[]{{\includegraphics[width=0.23\columnwidth]{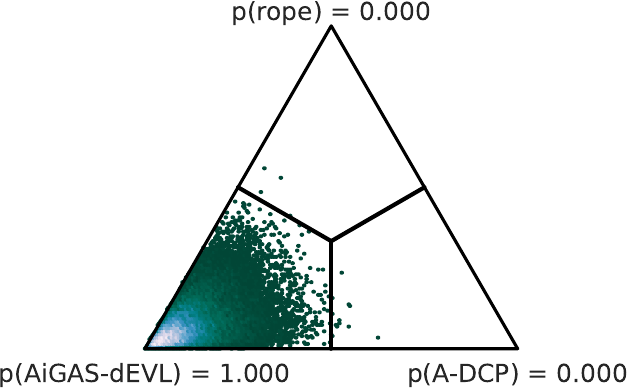}}}
	\hfill
	\subfloat[]{{\includegraphics[width=0.23\columnwidth]{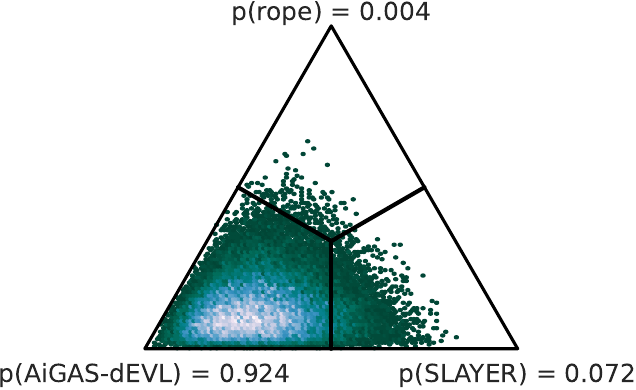}}}
	\newline
	\subfloat[]{{\includegraphics[width=0.23\columnwidth]{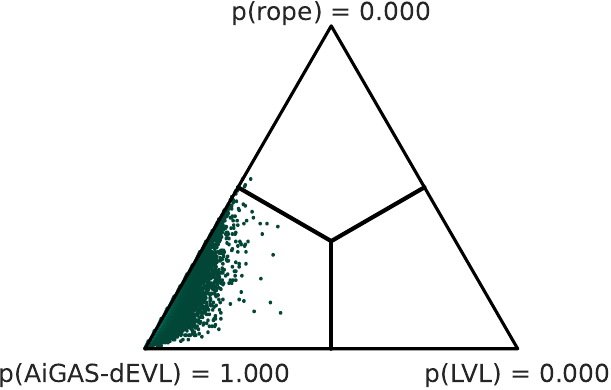}}}%
	\hfill
	\subfloat[]{{\includegraphics[width=0.23\columnwidth]{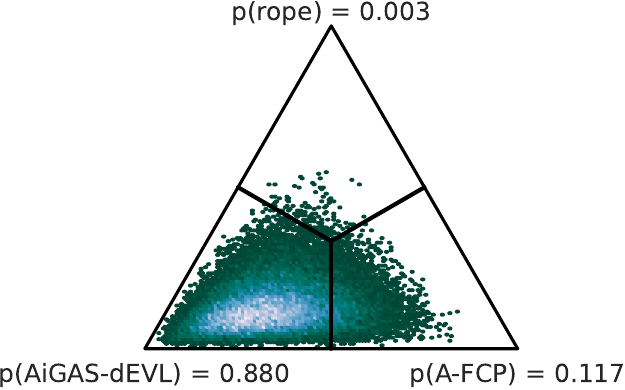}}}%
	\hfill
	\subfloat[]{{\includegraphics[width=0.23\columnwidth]{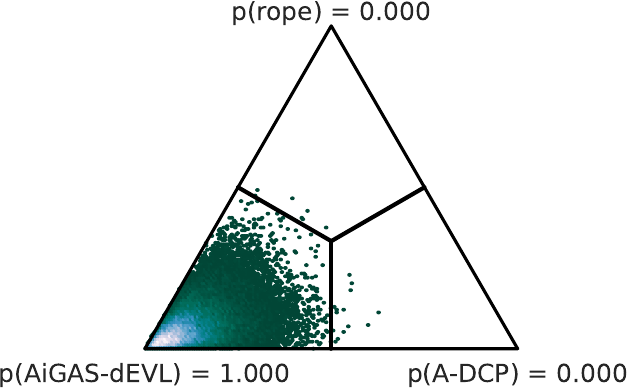}}}
	\hfill
	\subfloat[]{{\includegraphics[width=0.23\columnwidth]{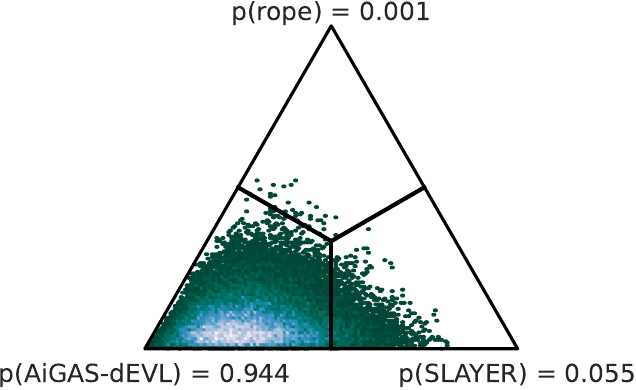}}}
	\caption{Bayesian posterior plots in barycentric coordinates comparing the differences in prequential error between \textbf{AiGAS-dEVL} and \textbf{LVL} (a, e), \textbf{A-FCP} (b, f), \textbf{A-DCP} (c, g), and \textbf{SLAYER} (d, h), for values of the rope parameter equal to 0.2 (top row) and 0.1 (bottom row).}
	\label{fig:bayesian}%
	
\end{figure}

To this end, the plots nested in Figure \ref{fig:bayesian} show the results of a Bayesian analysis of the distribution of the differences between \textbf{LVL}, \textbf{CMP}, \textbf{A-FCP}, \textbf{A-DCP}, \textbf{SLAYER} and the proposed \textbf{AiGAS-dEVL}. Bayesian posterior plots in barycentric coordinates permit to visually inspect probability distributions over three or more categories. In this representation, each vertex of a triangle corresponds to a hypothesis, and points within the triangle represent probability distributions over these categories. The position of a point is determined by its barycentric coordinates, which sum to 1 and indicate the relative probabilities of each category. The barycentric representation allows for easy interpretation of the most probable category (the closest vertex), the uncertainty in the estimate (distance from vertices), and the relative probabilities between categories (proximity to edges or vertices). In the context of Figure \ref{fig:bayesian}, we focus on the analysis of differences of the prequential error between \textbf{AiGAS-dEVL} and the approaches mentioned above, so that two of the regions in the triangle correspond to the hypothesis that one approach outperforms the other under comparison. The third region denotes statistical equivalence of the performance of the compared approaches determined by a parameter called \emph{rope}, which establishes the range of difference values that are considered practically equivalent to a specific value This parameter is given in the same units as the performance difference being modeled, hence it is interpretable and can be fixed depending on the purposes of the analysis.

Plots nested in Figure \ref{fig:bayesian} expose that \textbf{AiGAS-dEVL} clearly dominates the benchmark with statistical significance for rope equal to $0.2$ (top row) and $0.1$ (bottom row). Differences in terms of mean and standard deviation noted in Table \ref{tab:results} reflect on these plots as well, with the cases comparing \textbf{AiGAS-dEVL} to \textbf{SLAYER} and \textbf{A-FCP} being the only ones in which the posterior probability spans the three regions of the barycentric triangle. As expected, the posterior probability distribution in each of these regions become more separate apart from each other when the rope parameter is lower. 

\begin{tcolorbox}[notitle,boxrule=0pt,colback=gray!10,colframe=gray!20]
In response to RQ1, Tables \ref{tab:results} and \ref{tab:f1}, together with Figure \ref{fig:bayesian}, confirm that the proposed AiGAS-dEVL establishes a new performance landmark in modeling drifting data streams under extreme verification latency.
\end{tcolorbox}

\subsection*{RQ2: Do the points of improvement identified previously reflect on the performance of AiGAS-dEVL over time under different drift dynamics?}

To give an informed response to this second question, we proceed by analyzing how the performance scores evolve over time for different datasets. We focus on some of them that besides linking closely to the motivation for the research question, correspond to drift dynamics that can be inspected visually and interpreted geometrically, so that the performance decays can be attributed to non-rigid and/or non-smooth transformations, for which the projection in use within the naive AiGAS-dEVL design (line \textbf{13} in Algorithm 1) may not suffice.

\begin{figure}[h!]
	\begin{tabular}{c}
	\includegraphics[width=\columnwidth]{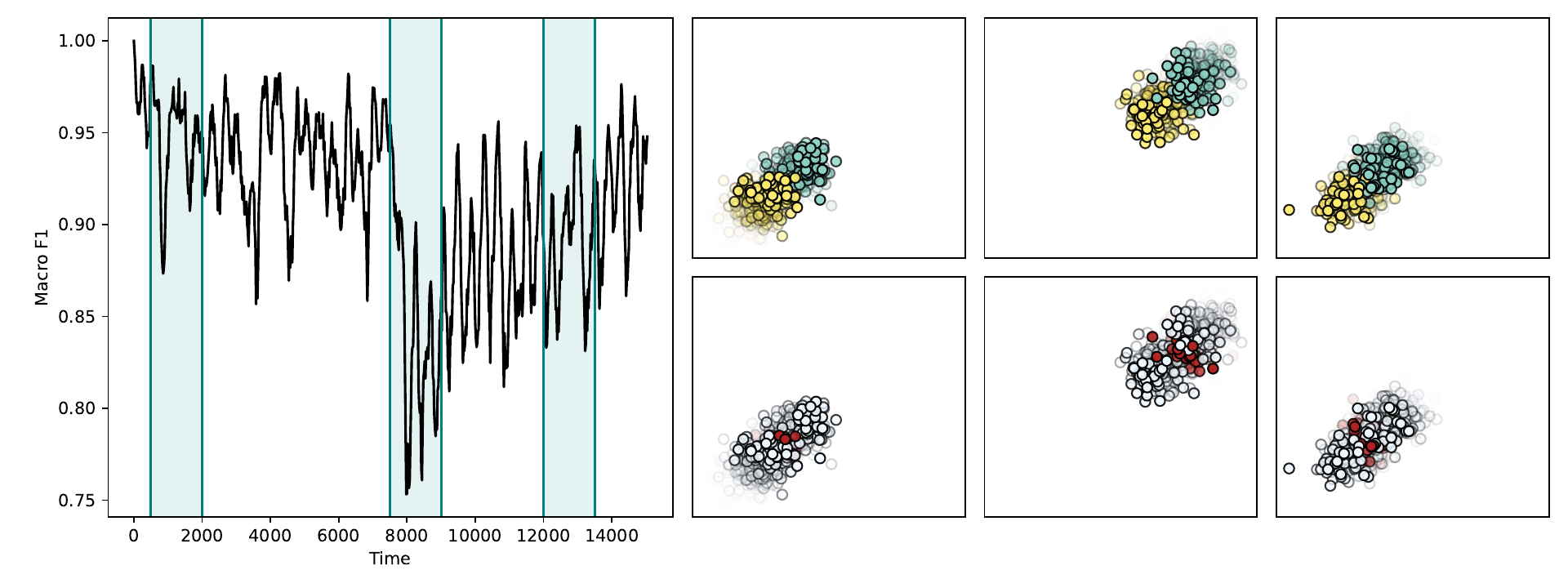}\vspace{-4mm}\\	
	(a) \texttt{2CDT} dataset.\\
	\includegraphics[width=\columnwidth]{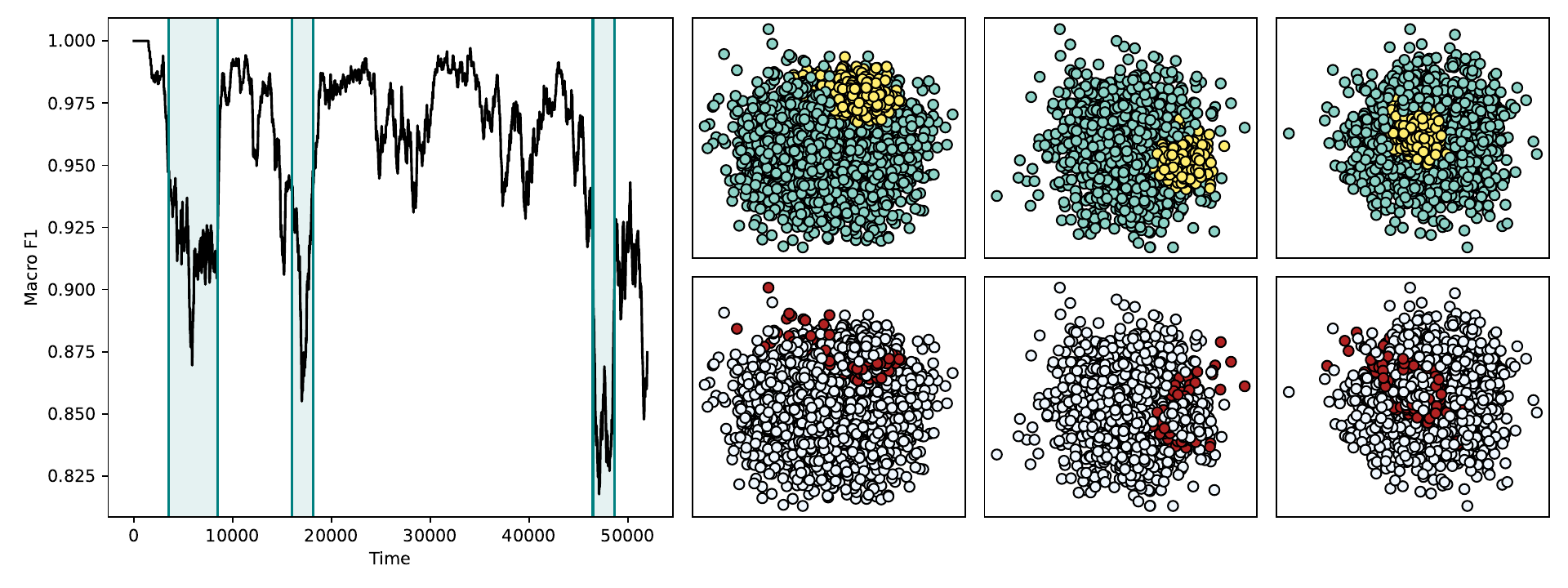}\vspace{-4mm}\\	
	(b) \texttt{1CSURR} dataset.\\
	\includegraphics[width=\columnwidth]{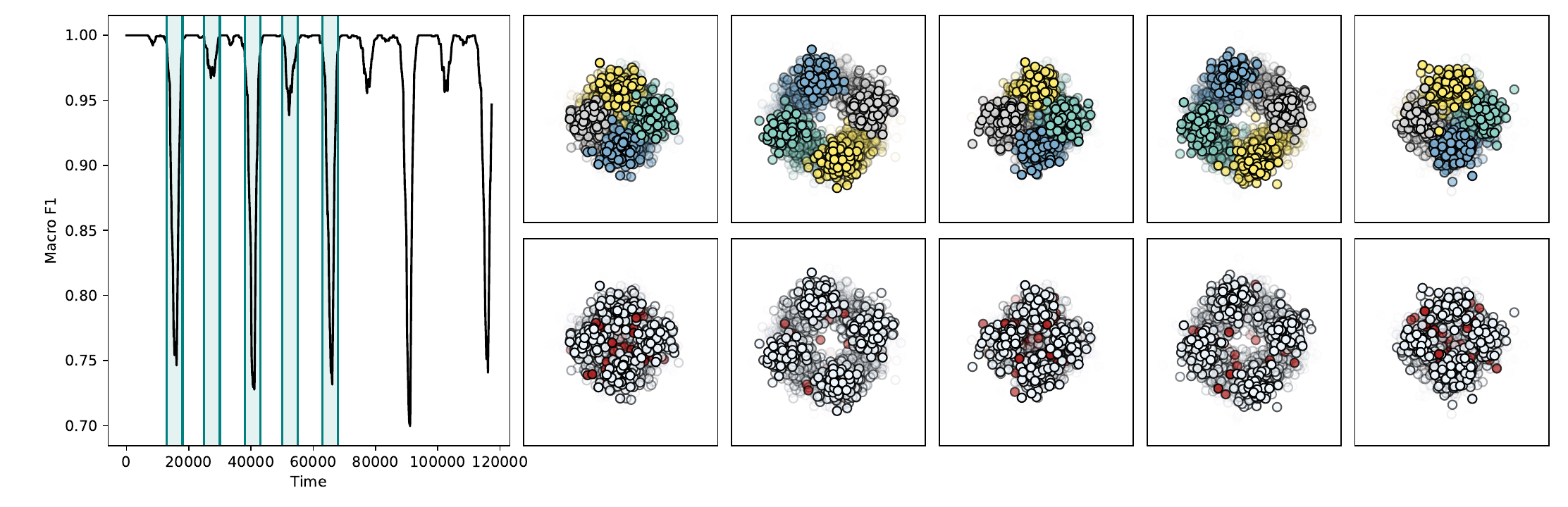}\vspace{-4mm}\\	
	(a) \texttt{4CRE-V1} dataset.\\
	\end{tabular}
	\caption{Evolution of the macro F1 score over time (on the left) and detail on the predictions (subplots on the right) for several datasets considered in the benchmark. Subplots on the right denote the predictions for every sample received during the periods highlighted in the evolution of the macro F1 score depicted on the left. Subplots in the bottom row identify streaming data instances that have been misclassified by the model, marking them in red.}
	\label{fig:4crev1_2CDT_1CSURR}
\end{figure}

To this end, we focus first on the plots included in Figure \ref{fig:4crev1_2CDT_1CSURR}, which depict the evolution of the macro F1 score over time for \texttt{2CDT} (Figure \ref{fig:4crev1_2CDT_1CSURR}.a), \texttt{1CSURR} (Figure \ref{fig:4crev1_2CDT_1CSURR}.b) and \texttt{4CRE-V1} (Figure \ref{fig:4crev1_2CDT_1CSURR}.c) datasets. Points within the depicted evolution of this performance metric have been via a sliding window of $B$ samples, with an overlap of $0.2B$ points between successive windows. On the right hand of every nested plot, we depict two rows of scatter plots. The ones in the top row denote the samples collected during the stream periods highlighted in the plot on the right, in temporal order and using color markers to denote the predicted class. The subplots in the bottom row can be interpreted likewise, but here red markers denote those samples whose classes predicted by AiGAS-dEVL are wrong (white otherwise). 

In response to RQ2, these plots expose several directions in which AiGAS-dEVL can be improved even further. To begin with, the analysis of limitations made in Subsection \ref{sec:complexity_adv_lim} anticipated a strong dependence of the algorithm on the model assumed for characterizing the drift dynamics over time, which was realized by a rigid point set registration between the distribution of GNG nodes corresponding to successive stream batches. Then, AiGAS-dEVL assumes a smooth continuity of such modeled drift dynamics, so that the new distribution of nodes is projected to estimate the region in the feature space where they will reside upon the reception of a new stream. This assumption does not hold when the trajectory of concepts during the unsupervised period of the stream changes suddenly, breaking the assumption of its smoothness and/or the rigid nature of the geometrical correspondence between GNG nodes. This can be noted in Figures \ref{fig:4crev1_2CDT_1CSURR}.a and \ref{fig:4crev1_2CDT_1CSURR}.b: 
\begin{itemize}[leftmargin=*]
	\item In the case of the \texttt{2CDT} dataset (Figure \ref{fig:4crev1_2CDT_1CSURR}.a), every class is characterizing by a single concept, which together delineate a linear drift trajectory from the $(min,min)$ (lower left corner) to the $(max,max)$ point (upper right corner) in the feature space. The performance is noted to decay around the middle of the unsupervised period of the stream, which coincides with the moment at which the two concepts suddenly start describing the same trajectory in the opposite direction. This sharp change yields a temporary mismatch between the estimated trajectory of the GNG nodes and the true distribution of classes in the feature space, giving rise to an increased number of wrongly classified stream instances. Clearly, the impact of this mismatch on the overall classification performance depends on whether the wrongly projected GNG nodes give rise to a $\textup{NN}$ decision boundary that does not follow the true distribution of classes in the stream. This is indeed what occurs in the \texttt{2CDT} dataset, as opposed to other datasets in which the mismatch does not affect the separability between classes (e.g. \texttt{1CHT}, which features a rectilinear drift with a sharp direction change, but only undergone by one of the concepts).
	
	\item A similar statement holds for the \texttt{1CSURR} dataset (Figure \ref{fig:4crev1_2CDT_1CSURR}.b), in which only the concept describing one of the two classes moves relative to the other. This concept evolves by traversing  the four corners of the feature space region over which both classes are distributed. It can be noticed that the performance decays when the evolving concept changes its direction to advance towards the next corner.

\end{itemize}

Another shortcoming identified in our discussion about limitations is the lack of memory in the assignment of labels to the GNG nodes computed for every new batch in the stream (line \textbf{11} in Algorithm 1). This can be detrimental in those cases where the drift dynamics make the distribution of several classes coexist in the same feature region. This effect is exemplified by the \texttt{4CRE-V1} dataset (Figure \ref{fig:4crev1_2CDT_1CSURR}.c), in which performance decays noticeably every time the concepts of the four classes collide with each other (as shown by the first, third and fifth regions depicted on the right subplots). A second smaller performance degradation occurs when the concepts, which undergo a continuous circular movement around the center of the region they occupy, expand and shrink suddenly along the radial axis. As mentioned before, sudden changes in the drift dynamics can affect the accuracy of predictions made by AiGAS-dEVL. However, in this case it can successfully recover the correct mapping between labels and evolving concepts, adapting better than other methods considered in the benchmark.

\begin{figure}[h!]
	\centering
	\begin{tabular}{c}
		\includegraphics[width=0.95\columnwidth]{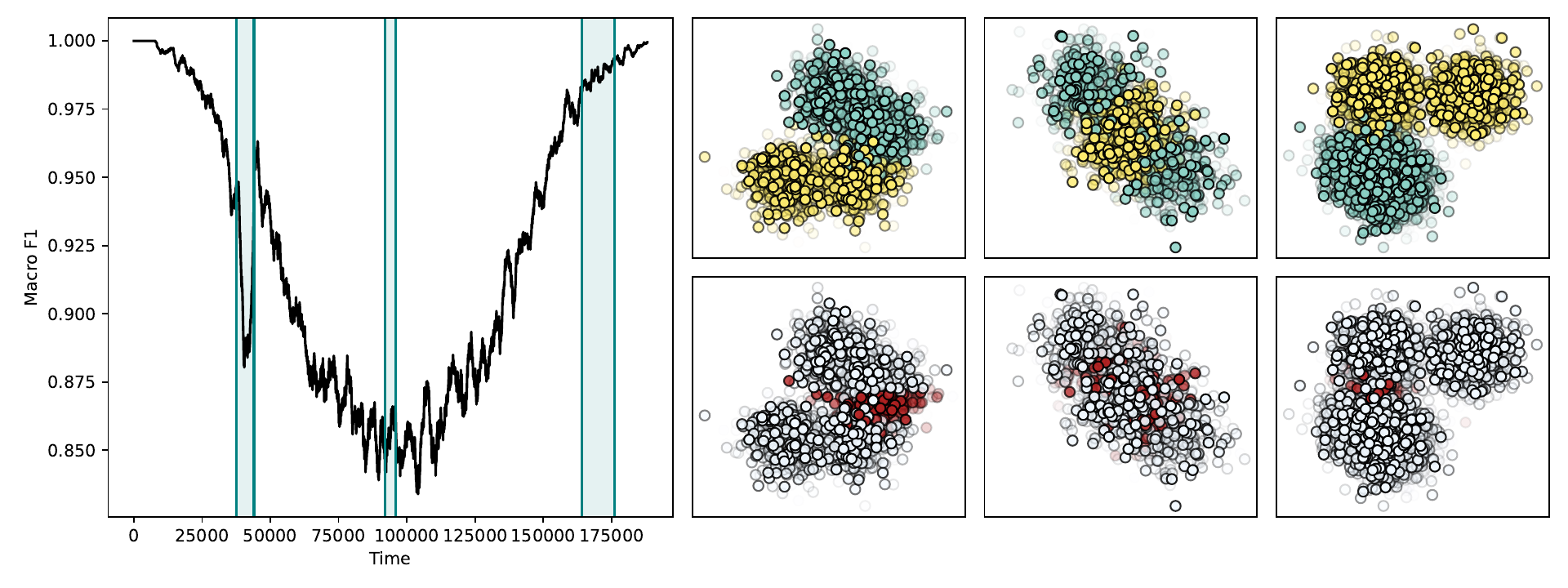}\vspace{-4mm}\\	
		(a) \texttt{MG2C2D} dataset.\\
		\includegraphics[width=0.95\columnwidth]{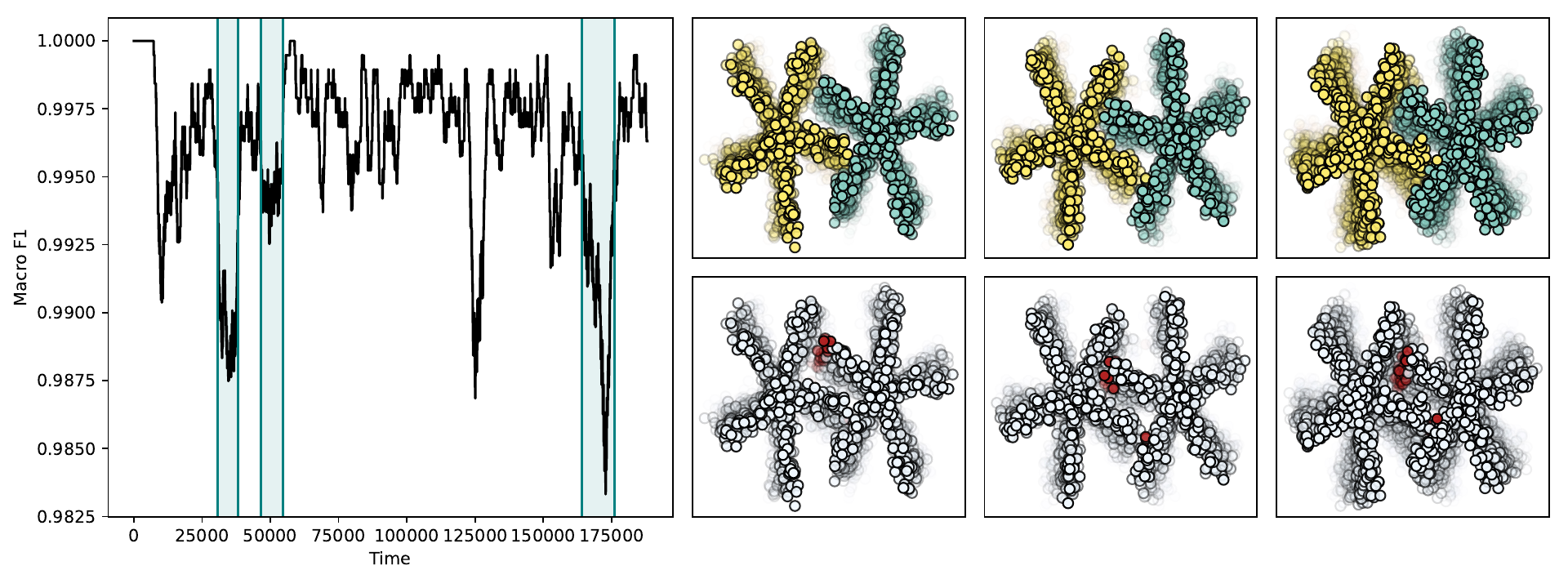}
		\vspace{-4mm}\\	
		(b) \texttt{GEARS} dataset.\\
		\end{tabular}
	\caption{Evolution of the macro F1 score over time (on the left) and detail on the predictions (subplots on the right) for several datasets considered in the benchmark. Similar interpretation as that detailed in the caption of Figure \ref{fig:4crev1_2CDT_1CSURR}.}
	\label{fig:MG2C2D_GEARS}
\end{figure}

We follow our discussion around RQ2 with the plots in Figure \ref{fig:MG2C2D_GEARS}, which depict similar plots to the previous ones, yet for the \texttt{MG2C2D} (Figure \ref{fig:MG2C2D_GEARS}.a) and \texttt{GEARS} (Figure \ref{fig:MG2C2D_GEARS}.b) datasets. The first case exemplifies the efficacy of AiGAS-dEVL to faithfully track the correspondence between GNG nodes and labels in a stream comprising two classes and several intertwined concepts per class. A performance degradation occurs due to the proximity between concepts in several parts of the stream, mainly due to the proximity between concepts of different classes and the corner cases in-between. A remarkable drop when concepts belonging to different classes coexist in the same feature space (first depicted region). However, once such concepts separate from each other, the point set matching and projection functionalities included at the algorithmic core of AiGAS-dEVL excel at recovering the performance level prior to the collision between concepts. The plots in Figure \ref{fig:MG2C2D_GEARS}.b corresponding to the \texttt{GEARS} dataset showcase the benefits of using GNG for the characterization of complex data that evolve over time. Indeed, the helix-shaped rotating data distributions of both classes in this dataset benefit from the adaptability of GNG, only failing to predict accurately instances in the boundary between both classes, and ultimately achieving notably superior performance metrics than its counterparts in Tables \ref{tab:results} and \ref{tab:f1}.
\begin{tcolorbox}[notitle,boxrule=0pt,colback=gray!10,colframe=gray!20]
	In response to RQ2, our discussion around Figures \ref{fig:4crev1_2CDT_1CSURR} and \ref{fig:MG2C2D_GEARS} concludes that the evolution of performance along time is in close match with the benefits and points of improvement identified in Section \ref{sec:complexity_adv_lim}, including the need for tailoring the selection of the projection to the drift dynamics of the stream, the effect of colliding concepts on the separability between classes, and the adaptability of GNG to complex data distributions.
\end{tcolorbox}

\section{Conclusions and Future Research Lines}\label{sec:conc}

This work has focused on the problem of dealing with extreme verification latency when modeling drifting data streams (dEVL). Under this assumption, concepts to be modeled inside the streaming data are non-stationary and hence evolve over time, demanding functionalities in the modeling approach to accommodate such changes in its captured knowledge. The lack of annotation exacerbates the difficulty of adapting to concept drift even further, as there is no supervisory signal from the stream to trace the correspondence between concepts and classes over time. 

Our work has tackled this problem from a new perspective by proposing an original modeling approach coined as AiGAS-dEVL. Our proposal embraces GNG at its heart to extract prototypes (\emph{nodes}) that characterize the data distribution over time. GNG is complemented with a matching between the GNG nodes extracted from consecutive batches, as well as with a projective mapping of such nodes based on the estimated drift dynamics of the stream. This projection allows predicting samples within the next batch more accurately, especially in the case of slowly evolving drifts. Furthermore, the overall combination of these algorithmic ingredients permit to overcome complex circumstances in dEVL scenarios where other approaches have traditionally struggled, such as the entanglement of concepts belonging to different classes in the feature space, or concepts with complex, non-corpuscular shapes. Besides the mathematical formalization of the proposed AiGAS-dEVL model, we have also critically examined its benefits and points of improvement, the latter aimed to highlight the plethora of research opportunities unleashed by this new approach to this practical problem.

We have evaluated and compared the performance of AiGAS-dEVL over a benchmark of datasets widely used in the dEVL literature. On one hand, we have confirmed that AiGAS-dEVL outperforms the rest of baselines in the benchmark with statistical significance. On the other hand, we have verified that the benefits and points of improvement identified from the algorithmic design of the model reflect on the behavior of its performance over time, connecting the distribution of classes and the drift dynamics of every dataset in the benchmark with our thoughts offered previously in this regard. Overall, we conclude that despite the excellent performance reported in this study, the AiGAS-dEVL model presented in this manuscript should be conceived as a realization of a more general design template, in which the classifier, the matching algorithm between GNG nodes and/or their projection should be chosen as per the traits of the data stream under consideration and the dynamics of the drift.

In line with the overarching conclusion drawn from our work, we envision several research directions to be tackled in the future, with the goal of overcoming some of the shortcomings identified in our critical analysis. To begin with, characterizing the trajectory of the data distribution induced by the presence of concept drift can be enhanced by exploring other elements from the point set registration research area, including recent learning-based methods that approximate non-rigid transformations \cite{yuan2023non}. Furthermore, we plan to enhance the assignment of labels to the newly discovered GNG nodes by devising inertial algorithms that make the assignment depend not only on the GNG nodes belonging to the previous batch, but also on the assignment and distances of previous batches. A balance must be met in this regard between the stability and plasticity of the assignment: stability is needed for the assignment to be robust against small drifts and noise in the dataset propagating to the update of the GNG nodes, whereas plasticity could favor the adaptability of AiGAS-dEVL to sharply changing concepts. Endowing our approach with the capability to tune this trade-off autonomously over time will be among our research priorities in the near future.

\section*{Acknowledgments}

M. Arostegi, J. L. Lobo and J. Del Ser receive funding support from the FaRADAI project (ref. 101103386) granted by the European Commission under the European Defence Fund (EDF-2021-DIGIT-R). J. Del Ser also acknowledges funding from the Basque Government through the consolidated research group MATHMODE (ref. IT1456-22).

\bibliographystyle{model1-num-names}
\bibliography{biblio}

\end{document}